\documentclass{article}
\usepackage{ijcai23}

\usepackage{times}
\usepackage{soul}
\usepackage{url}
\usepackage[utf8]{inputenc}
\usepackage[small]{caption}
\usepackage{graphicx}
\usepackage{amsmath}
\usepackage{amsthm}
\usepackage{booktabs}
\usepackage{algorithm}
\usepackage[switch]{lineno}

\usepackage{amssymb}
\usepackage{nicefrac}       %
\usepackage{microtype}      %
\usepackage{pifont}
\usepackage{multirow}
\usepackage{verbatim}
\usepackage{color,xcolor}
\usepackage{enumitem}
\usepackage{tabulary,overpic}
\usepackage[caption=false]{subfig}
\usepackage{epstopdf}
\usepackage{algorithmicx}
\usepackage{algpseudocode}
\usepackage{enumerate}
\usepackage{bm}
\usepackage{colortbl}

\usepackage[hidelinks]{hyperref}

\definecolor{row}{RGB}{235, 245, 251}

\newcommand{\cmark}{\ding{51}}%
\newcommand{\xmark}{}%
\newcommand{\fref}[1]{Fig.~\ref{#1}}
\newcommand{\tref}[1]{Table~\ref{#1}}

\newcommand{\ch}{}
\newcommand{\tablestyle}[2]{\setlength{\tabcolsep}{#1}\renewcommand{\arraystretch}{#2}\centering\footnotesize}

\title{SWAT: Spatial Structure Within and Among Tokens}

\author{
    Kumara Kahatapitiya \And Michael S. Ryoo\\
    \affiliations
    Stony Brook University\\
    \emails
    \{kkahatapitiy, mryoo\}@cs.stonybrook.edu
}

\begin{document}

\maketitle

\begin{abstract}

Modeling visual data as tokens (i.e., image patches) using attention mechanisms, feed-forward networks or convolutions has been highly effective in recent years. Such methods usually have a common pipeline: a tokenization method, followed by a set of layers/blocks for information mixing, both within and among tokens. When image patches are converted into tokens, they are often flattened, discarding the spatial structure within each patch. As a result, any processing that follows (eg: multi-head self-attention) may fail to recover and/or benefit from such information. In this paper, we argue that models can have significant gains when spatial structure is preserved during tokenization, and is explicitly used during the mixing stage. We propose two key contributions: (1) Structure-aware Tokenization and, (2) Structure-aware Mixing, both of which can be combined with existing models with minimal effort. We introduce a family of models (\textbf{SWAT}), showing improvements over the likes of DeiT, MLP-Mixer and Swin Transformer, across multiple benchmarks including ImageNet classification and ADE20K segmentation. Our code is available at \href{https://github.com/kkahatapitiya/SWAT}{\texttt{github.com/kkahatapitiya/SWAT}}.

\end{abstract}
\section{Introduction}

Convolutional architectures (CNNs) \cite{he2016deep} have been dominant in computer vision for a while now. When they were first introduced for large-scale training in image domain, their benefits were quickly realized over Multi-layer Perceptrons (MLPs). In addition to efficient weight sharing, the inductive bias generated by exploring the local structure in images was one of the key factors for its success \cite{lecun2015deep}. In language domain however, CNNs were less effective due to lack of such strong local structure. Consequently, attention mechanisms emerged dominant, exploring long-range relationships and modeling language as a sequence \cite{dauphin2017language}. 
More recently, attention models-- specifically Transformers \cite{vaswani2017attention}, have been extended to represent visual data \cite{dosovitskiy2020image}, with the key concept of tokenizing an input image to create a sequence (or a set), often discarding their structure. Within a short period of time, such \textit{token-based models} (i.e., class of models such as ViTs \cite{dosovitskiy2020image} and MLP-Mixers \cite{tolstikhin2021mixer}) have outperformed CNNs on most visual tasks. However, we ask, could the spatial structure-- when preserved, benefit token-based models and further improve their performance? 

\begin{figure*}[t]
	\centering
	\includegraphics[width=.95\linewidth]{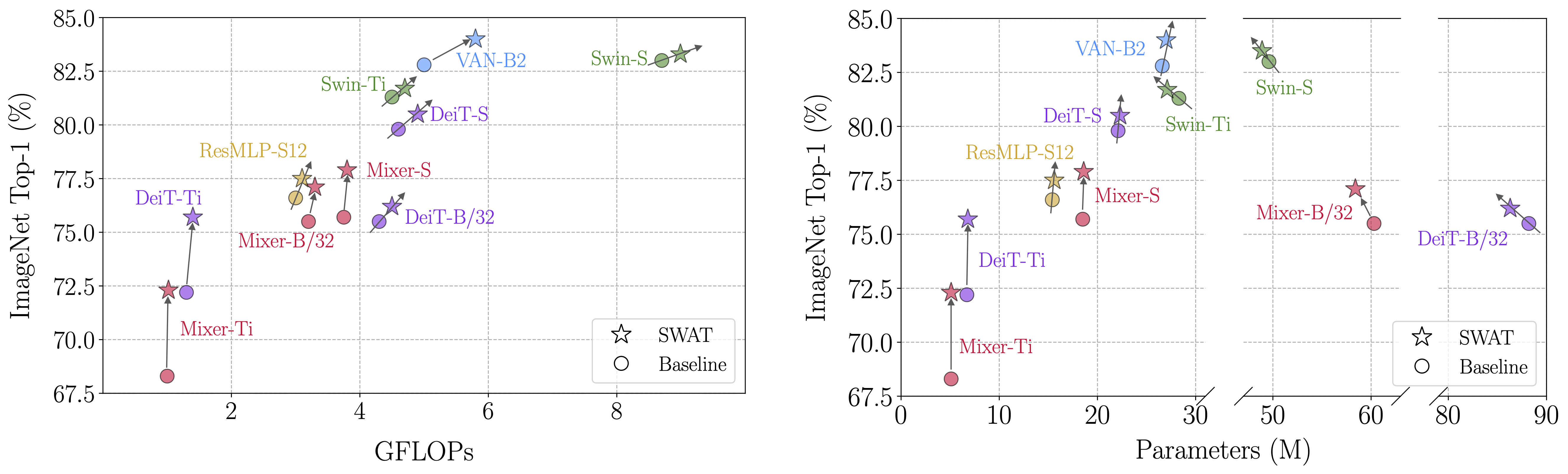}
	\caption{\textbf{Performance vs. Complexity} on ImageNet-1K \protect\cite{deng2009imagenet}. We implement our proposed  (1) \textit{Structure-aware Tokenization}, and (2) \textit{Structure-aware Mixing} in common token-based architectures including DeiT \protect\cite{touvron2021deit}, Swin \protect\cite{liu2021swin}, MLP-Mixer \protect\cite{tolstikhin2021mixer}, ResMLP \protect\cite{touvron2021resmlp} and VAN \protect\cite{guo2022van}. The resulting family of \textbf{SWAT} models consistently outperform their counterparts, with minimal increase in complexity. We consider the system-agnostic metrics such as FLOPs and Parameters as the complexity measures here.}
	\label{fig:acc_vs_compute}
\end{figure*}

Token-based models in computer vision are rapidly evolving. From Vision Transformers \cite{dosovitskiy2020image} to MLP-Mixers \cite{tolstikhin2021mixer} and hybrid-architectures \cite{peng2021conformer,wu2021cvt}, intriguing concepts are being introduced and tested on tasks including classification \cite{dosovitskiy2020image,touvron2021deit,liu2021swin}, detection \cite{zhu2020deformable,dai2021up} and segmentation \cite{xie2021segformer,duke2021sstvos}, to name a few. All such models can be framed with two main components: (1) \textit{Tokenization}, which converts image patches into tokens, and (2) \textit{Mixing} (attention-based as in Multi-head Self Attention (MHSA), MLP-based or convolution-based), which shares information within and among tokens. In general, during tokenization, an image patch is directly mapped into a token, not preserving the spatial structure within a token.
After this mapping, models usually focus on global patterns among tokens, without capturing local spatial structure \textit{within} tokens.

Structure is an important cue in visual data. In images, 2D spatial structure preserves geometry and object-part relationships.
Simply put, structure gives meaning to visual data in human perspective. However, in machine perspective, if a jumbled set of image patches are tokenized and processed through a token-based model, it can give the same classification performance (as it is a set operator), even though the input is really meaningless to a human \cite{naseer2021intriguing}. This is in fact a drawback of token-based models (eg: can be prone to such an adversarial attack), which could be addressed by structure-aware modeling. Not only the structure among tokens, but also the structure within tokens is equally-important which is often discarded during tokenization. It is particularly beneficial to maintain the structure within tokens for fine-grained prediction tasks such as segmentation.

In this paper, we propose to preserve and make use of the spatial structure both within and among tokens. To do this we focus on two components: (1) \textit{Structure-aware Tokenization} and (2) \textit{Structure-aware Mixing}\footnote{Information sharing based on either attention (MHSA), MLPs or convolutions is commonly referred to as \textit{Mixing} in this paper.}, both of which can be adopted in existing token-based architectures with minimal effort. Our Structure-aware Tokenization converts image patches to tokens, but \textit{preserves the spatial structure within a patch as channel segments of the corresponding token}. Our Structure-aware Mixing benefits from the preserved structure by \textit{considering local neighborhoods both within and among tokens}, based on 2D convolutions. We also refer to this as token mixing with channel structure and channel mixing with token structure. With these two contributions, we introduce a family of models: \textbf{SWAT}, and compare against common baselines such as DeiT \cite{touvron2021deit}, Swin Transformer \cite{liu2021swin}, MLP-Mixer \cite{tolstikhin2021mixer}, ResMLP \cite{touvron2021resmlp} and VAN \cite{guo2022van}. Our models show consistent improvements over baseline models on multiple benchmarks including ImageNet-1K \cite{deng2009imagenet} classification and ADE20K \cite{zhou2019ade20k} semantic segmentation. We further visualize fine-grained attention patterns captured by our structure-aware modeling. Performance gains on ImageNet-1K classification against complexity (measured by system-agnostic metrics such as FLOPs and Parameters) are shown in \fref{fig:acc_vs_compute}.

\section{Related Work}
\label{sec:related}

\paragraph{Token-based models:} Transformer architectures from language domain \cite{vaswani2017attention,devlin2018bert} have been recently adopted to visual data in the seminal work ViT \cite{dosovitskiy2020image}. 
Even though attention mechanisms already existed in computer vision \cite{wang2018non,zhao2020exploring}, their true potential was realized when introduced with tokenization. Since then, a variety of token-based models have been introduced, some with the use of MLPs \cite{tolstikhin2021mixer,touvron2021resmlp} or convolutions \cite{trockman2022patches,liu2022convnet}.
DeiT \cite{touvron2021deit} introduces an efficient training recipe, and \cite{caron2021emerging,ranasinghe2022self} use self-supervision. 
Swin Transformer \cite{liu2021swin} introduces attention within shifted-windows, while downsampling progressively similar to \cite{heo2021rethinking,wang2021pyramid,fan2021multiscale}.
Another direction explores efficiency of such models \cite{zhai2021attention,bello2021lambdanetworks,graham2021levit,tang2021sparse,yue2021vision,ryoo2021tokenlearner}.

\paragraph{Token adoption in vision tasks:} Token-based models are already applied in most vision applications, including classification \cite{touvron2021deit,liu2021swin}, object detection \cite{zhu2020deformable,carion2020end}, segmentation \cite{xie2021segformer,duke2021sstvos}, image generation \cite{cao2021video,esser2021taming}, video understanding \cite{nagrani2021attention,fan2021multiscale,arnab2021vivit,dai2022ms}, dense prediction \cite{yang2021transformer,ranftl2021vision}, point clouds processing \cite{zhao2021point,guo2021pct} and reinforcement learning \cite{chen2021decision,shang2021starformer}.

\begin{figure*}[t!]
	\centering
	\includegraphics[width=0.9\textwidth]{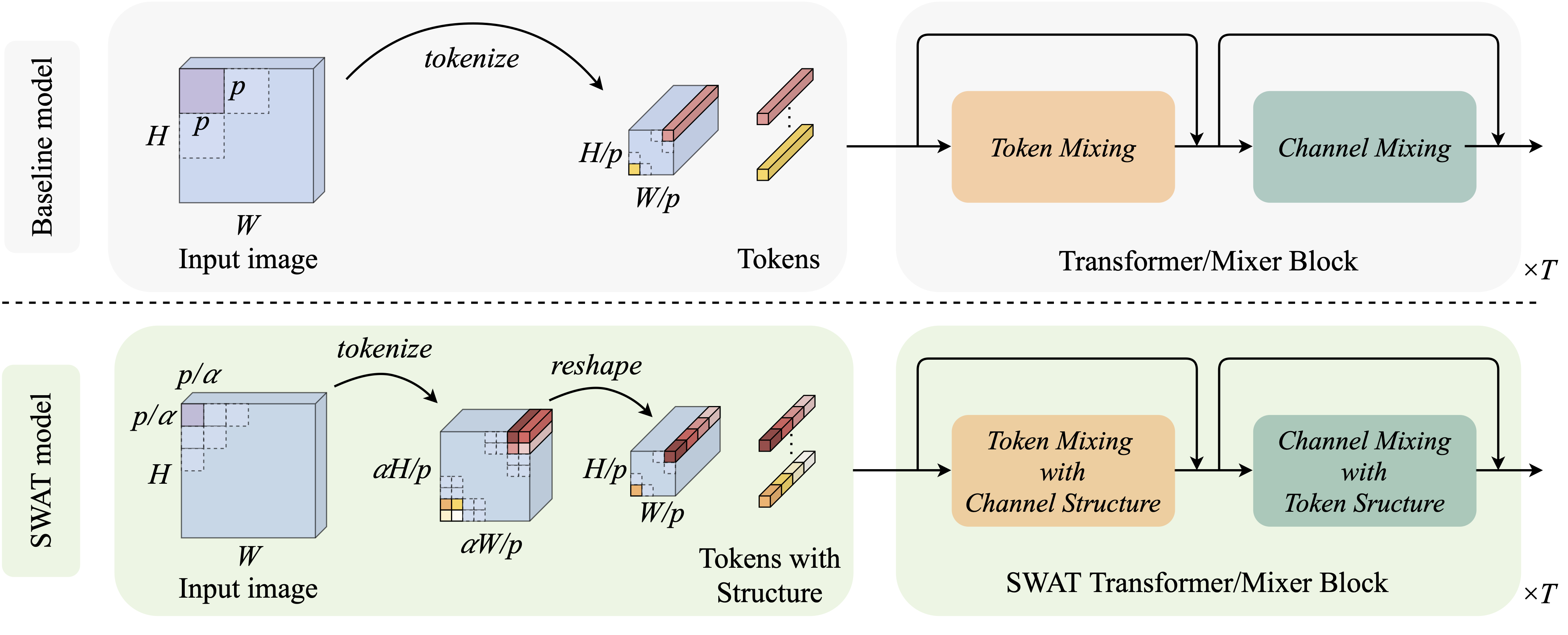}
	\caption{\textbf{SWAT Overview:} We show the architecture of SWAT (bottom) and a baseline model (top) in this figure. We propose two main contributions: (1) Structure-aware Tokenization and, (2) Structure-aware Mixing, which can be applied to common Transformer or Mixer architectures with minimal effort. Structure-aware tokenization preserves the spatial structure within a token, as channel segments. Simply put, first, we tokenize with a patch size $(p/\alpha\times p/\alpha)$ instead of $(p\times p)$, resulting in $\times\alpha^2$ more intermediate tokens. Next, we restructure $\alpha\times\alpha$ neighboring tokens into one token (concatenating in channel dimension), which gives the same number of tokens (and the channel dimension) into  Mixing operations, ensuring no additional cost in the downstream. However, now, this newly-preserved structure within tokens can be explored for better downstream processing, which was very limited previously. More on how we use the structure in Mixing is shown in \fref{fig:swat_deit} and \fref{fig:swat_mixer}.}
	\label{fig:overview}
\end{figure*}

\paragraph{Structure with token-based models:} Some prior work in token-based models have explored structure, using hybrid architectures with convolutions \cite{xiao2021early,peng2021conformer,d2021convit}. 
A structure-based grouping method is proposed in T2T-ViT \mbox{\cite{yuan2021tokens}}. With a complementary motivation to ours, TNT \cite{han2021transformer} and NesT \cite{zhang2021aggregating} both consider a sub-token structure within tokens, but introduce additional tokens and become heavier with extra processing. \cite{yuan2021incorporating} has similarities with our channel mixing with token structure. 
Models such as ConvMixer \cite{trockman2022patches}, ConvNeXt \cite{liu2022convnet} and VAN \cite{guo2022van} also consider a convolutional design as ours (w/ Pointwise Conv and Depthwise Conv). However, they only consider structure \textit{among} tokens, not structure \textit{within} tokens. 
To our knowledge, this is the first work to preserve structure within tokens, without extra tokens or processing, i.e., with a minimal change in footprint.

\section{\hspace{-1mm}Spatial Structure Within and Among Tokens}
\label{sec:method}

In SWAT family of models, we explore the benefits of preserving spatial structure not only among tokens, but within tokens as well. To do this with a general framework, we consider all token-based models (eg: ViTs \cite{dosovitskiy2020image}, Mixers \cite{tolstikhin2021mixer}) as a unified architecture, which consists of two main components: (1) \textit{Tokenization}, for converting image patches into tokens, and, (2) \textit{Mixing}, for sharing information within and among tokens. Mixing can mean either the use of Multi-Layer Perceptron (MLP), Multi-Headed Self-Attention (MHSA) or convolution for information sharing. In this framework, we suggest improvements to both Tokenization and Mixing. \ch{When these components are adopted \textit{together} in a network, it can preserve and utilize the spatial structure.} Namely, we introduce \textit{Structure-aware Tokenization} and \textit{Structure-aware Mixing}, which we describe below in detail.

\subsection{Structure-aware Tokenization}

Here, we propose to preserve the spatial structure within tokens, not imposing any additional burden on downstream processing. The idea is to keep spatial information within tokens separated as its channel segments, so that the `mixing' component can later take advantage of it. In general, image patches are converted into tokens by sliding a large convolutional kernel with a stride (eg: a $16\times16$ kernel with a stride of 16), which extracts a set of tokens. In such a setting, all the spatial information within a patch is directly fused into the channels of the corresponding token, losing the explicit structure in the process. In our method, we replace this direct fusion, retaining structural information within tokens.

More concretely, let us consider an input image of size $H\times W\times 3$, and a baseline tokenizer which converts image patches into tokens by extracting non-overlapping patches of size $p\times p$. This is usually implemented as a convolutional layer with $C$ kernels of size $(p\times p)$, applied at a stride of $p$. The output here will be an $H/p\times W/p$ 2D structure of tokens, which is reshaped to create a sequence of $HW/p^2$ tokens of embedding dimension $C$ (refer \fref{fig:overview} top). Even though these tokens are processed downstream as a sequence, they can be reshaped back into the original 2D structure of $H/p\times W/p$ whenever necessary. It has been observed that the tokens preserve this structure (\textit{among} tokens) through skip connections and positional encodings \cite{caron2021emerging,naseer2021intriguing}, 
even after a series of Mixing blocks. However, the structure within a $p\times p$ patch is irreversibly lost, i.e., although each token is a linear abstraction of $p\times p$ pixels, remapping the token back to its original $p\times p$ shape
 in subsequent layers is not directly feasible.

In contrast, the proposed tokenizer in SWAT retains the structure \textit{within} a token (refer \fref{fig:overview} bottom). We do this by first having $C/\alpha^2$ convolutional kernels of size $(p/\alpha\times p/\alpha)$ (where $\alpha>1$) and applying it with a stride of $p/\alpha$. The resulting intermediate set of tokens will have a 2D structure of $\alpha H/p\times \alpha W/p$ and a dimension of $C/\alpha^2$. Next, such  $\alpha\times\alpha$ neighboring tokens are reshaped into a single token (concatenating in the channel dimension), creating the same number of tokens $HW/p^2$ as the baseline, with the embedding dimension of $C$. By doing so, we now have an $\alpha\times\alpha$ 2D structure within each token-- as its channel segments, which can be preserved throughout downstream processing, by the same principles: skip connections and (optional) positional embeddings. Note that the SWAT tokenizer will have a fewer parameters, in fact,  $3Cp^2/\alpha^4$, compared to that of the baseline ($3Cp^2$), which can impair the learning capacity. To avoid this in practice, we use a bottleneck structure of multiple layers instead of a single convolution layer (still having the same downsampling factor of $1/\alpha$ as the baseline), which will enable the tokenizer to have an equivalent capacity, while introducing structure within tokens.

\subsection{Structure-aware Mixing}

To make use of the structured tokens (w/ spatial structure both within and among) generated by the SWAT tokenizer, we propose \textit{Structure-aware Mixing}.
The idea is straightforward: when we have such a 2D structure, the corresponding elements (either tokens or channels) will have the notion of neighboring elements in the 2D space, which gives an inductive bias that we can benefit from. Our approach uses this locality in a form of 2D convolutions, mixing information in a local region of elements, in addition to the usual global information sharing in Transformer/Mixer models. We present this idea in two parts: (1) \textit{Token Mixing with Channel Structure} and, (2) \textit{Channel Mixing with Token Structure}.

\subsubsection{Token Mixing with Channel Structure}
\vspace{2mm}

\begin{figure}[t]
	\centering
	\includegraphics[width=1\linewidth]{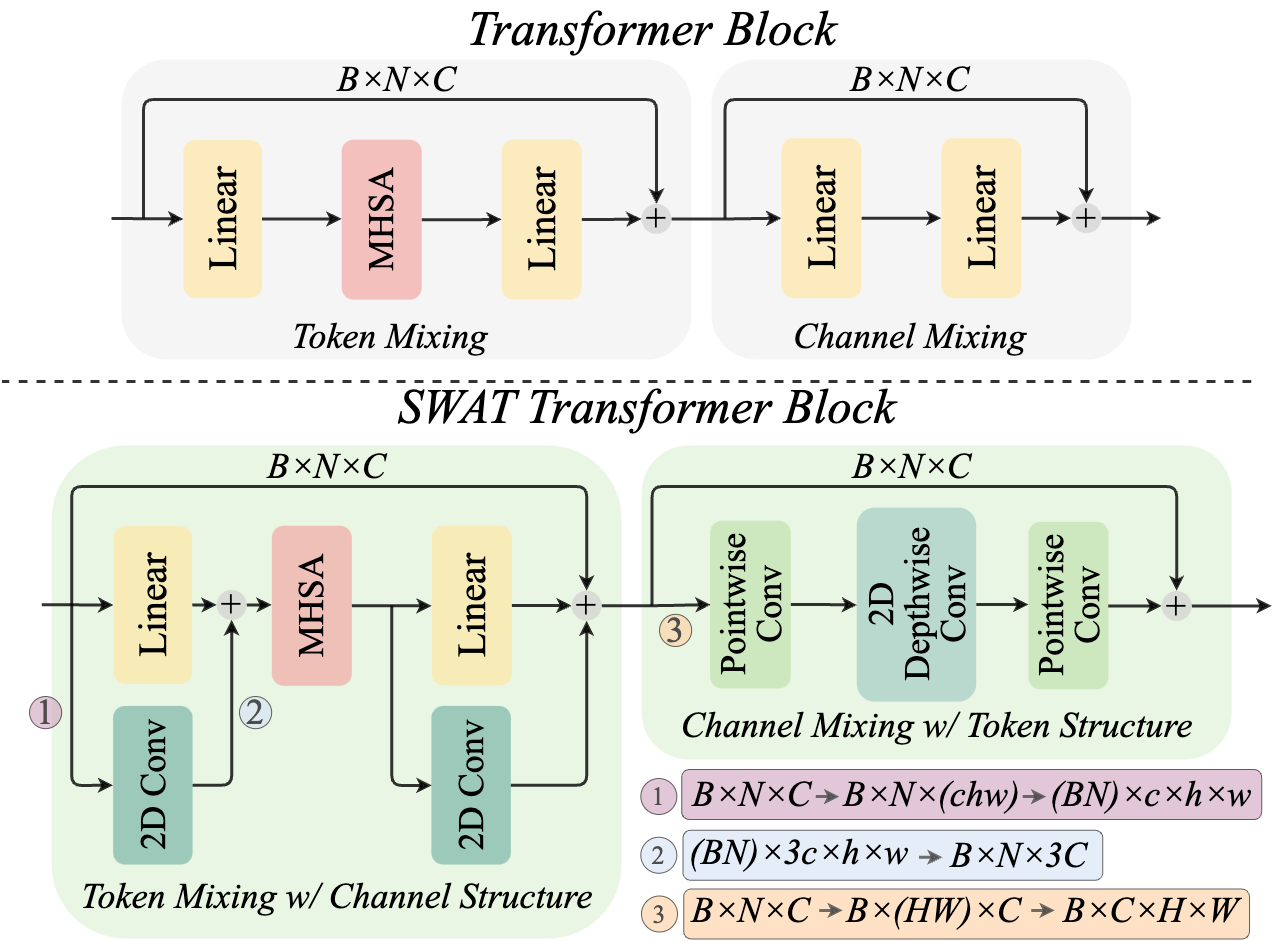}
	\caption{\textbf{SWAT Transformer Block:} We benefit from 2D channel structure (within tokens) in token mixing, and 2D token structure (among tokens) in channel mixing: (1) We insert a 2D Conv in-parallel to each Linear projection in attention (MHSA) block, applied on a reshaped input. It explicitly considers a structured local neighborhood \textit{within} a token during token mixing. (2) In channel mixing, we first replace Linear layers with Pointwise Conv as a design decision for implementation simplicity. Next, we insert a 2D Depthwise Conv to consider a structured local neighborhood \textit{among} tokens. Key tensor reshape operations are highlighted.}
	\label{fig:swat_deit}
\end{figure}

Token Mixing happens in different ways in Transformers \cite{dosovitskiy2020image,touvron2021deit} and Mixers \cite{tolstikhin2021mixer,touvron2021resmlp}. In Transformers, each token attends to every other token pairwise and dynamically (w/ input-dependent weights). In an attention block, a MHSA layer is sandwiched between two Linear projection layers. Here, by design, token mixing (i.e., information sharing among tokens) happens while also mixing channels. These Linear layers may reshuffle channels and waste our newly-introduced structure within tokens, as there is not even a skip-connection to save it. 
In contrast, in Mixers, token mixing is done with static relations (w/ learned weights), while not reshuffling channels. Simply put, tokens are mixed channel-wise, without damaging the structure within tokens. Therefore, we follow different designs in Transformers and Mixers to introduce our \textit{token mixing with channel structure}.

\begin{figure}[t]
	\centering
	\vspace{1.mm}
	\includegraphics[width=1\linewidth]{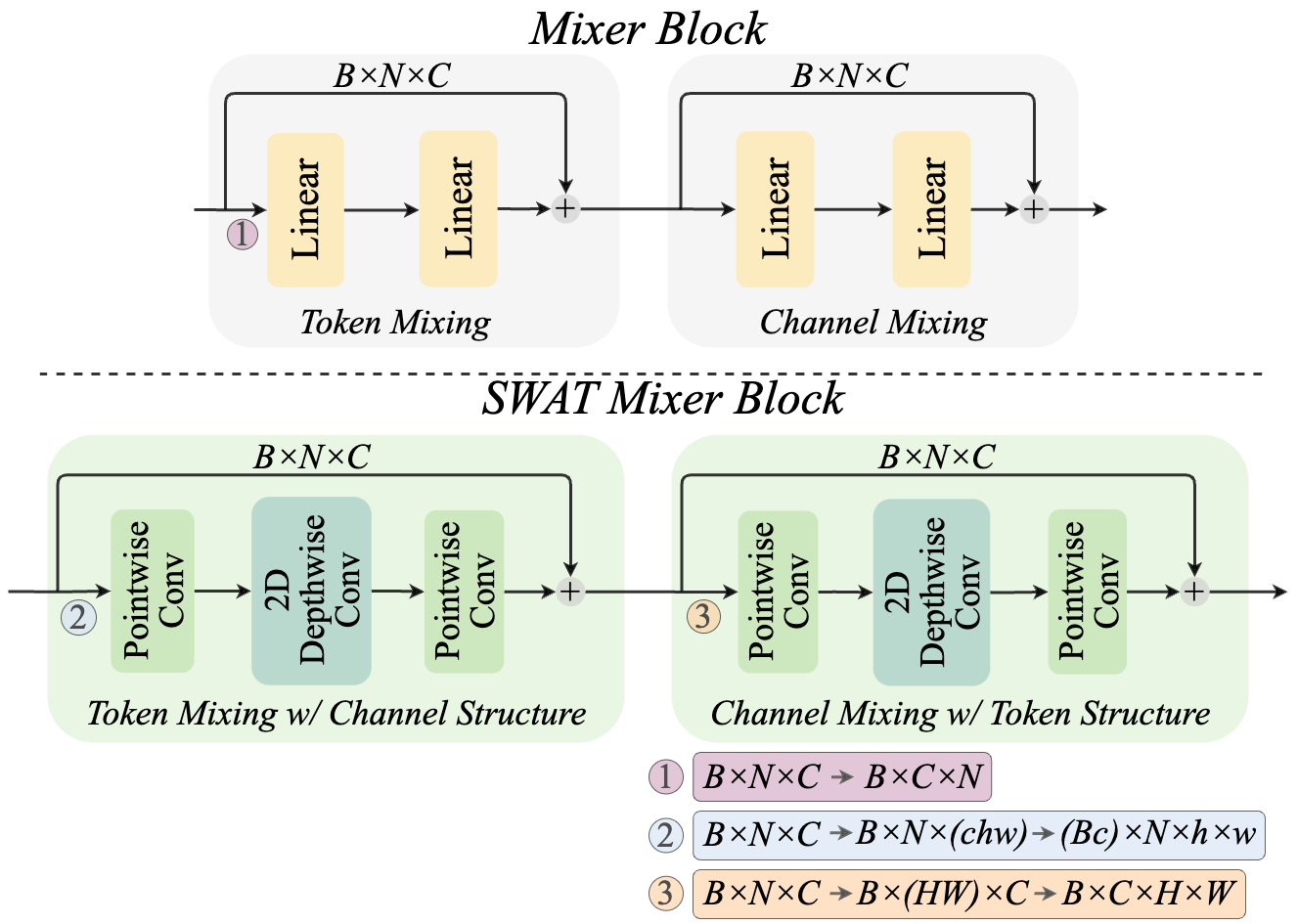}
	\vspace{-3mm}
	\caption{\textbf{SWAT Mixer Block:} In both token mixing and channel mixing, we first replace Linear layers with Pointwise Conv, applied on a reshaped input (eg: a Linear layer on a ($B\times N\times C$) shaped tensor equals to a Pointwise Conv on ($B\times C\times N$), in a PyTorch-like channel-first Conv implementation). This is a design decision for implementation simplicity, which makes no change in how an input is processed. Now, we can easily explore the 2D channel structure (\textit{within} tokens) in token mixing and, 2D token structure (\textit{among} tokens) in channel mixing, by inserting a 2D Depthwise Conv. Key tensor reshape operations are highlighted.}
	\label{fig:swat_mixer}
\end{figure}

\paragraph{Transformers:} We insert a 2D Conv in-parallel\footnote{Why in-parallel? To retain a capacity (params) similar to the baseline. Refer to Appendix for more details.} to the Linear layers before and after MHSA, to explore the channel structure (structure within tokens). See \fref{fig:swat_deit} bottom-left. After SWAT tokenizer, the channel dimension $C$ has an internal structure of $c\times h\times w$ (as in \fref{fig:overview}, with usual notation), which we use to reshape the input as,
\begin{equation*}
    B\times N\times C \rightarrow B\times N\times (chw) \rightarrow (BN)\times c\times h\times w.
\end{equation*}
\noindent Here, $B$ represents batch, $N$, num. of tokens and $C$, embedding dimension. When a 2D Conv is applied on this tensor\footnote{Here we consider a PyTorch-like channel-first implementation of convolution (eg: 2D Conv has an input shape of $B\times C\times H\times W$).}, it can mix channel information similar to a Linear layer, but also considering the inductive bias of channel structure.

\paragraph{Mixers:} In Mixers, we first replace the two Linear layers in token mixing with Pointwise $1\times1$ Conv. See \fref{fig:swat_mixer} bottom-left. We do this just to simplify the implementation, w/o changing the underlying operation (i.e., Linear = $1\times1$ Conv). Applying a Linear layer on a tensor of shape $B\times C\times N$ is the same as applying a Pointwise Conv on  a tensor $B\times N\times C$ (again, we consider a PyTorch-like implementation of channel-first Conv and channel-last Linear). Now, we can conveniently consider the 2D structure in channels (\textit{within} tokens). Next, we insert a 2D Depthwise Conv in-between the Pointwise Conv layers, applied on a reshaped input as,
\begin{equation*}
    B\times N\times C \rightarrow B\times N\times (chw) \rightarrow (Bc)\times N\times h\times w.
\end{equation*}
\noindent Altogether, this token mixing block now considers the channel structure (i.e., structure \textit{within} tokens).

\subsubsection{Channel Mixing with Token Structure}
\vspace{2mm}

Channel Mixing operation is the same for both Transformers and Mixers. In a baseline, two Linear layers are applied on an input tensor shaped as $B\times N\times C$ to mix channel information. In SWAT, we wish to do this while considering the token structure. Hence, we replace the two Linear layers with the same sandwich block: 2D Depthwise Conv in-between two Pointwise Conv, applied on an input reshaped as,
\begin{equation*}
    B\times N\times C \rightarrow B\times (HW)\times C \rightarrow B\times C\times H\times W.
\end{equation*}
\noindent See \fref{fig:swat_deit} or \fref{fig:swat_mixer} bottom-right. This channel mixing block now considers token structure (i.e., structure \textit{among} tokens).

\vspace{2mm}
Specific hyperparameter settings and ablations related to (1) newly-introduced structure within tokens, and (2) level of structure-awareness in mixing, are included in Appendix. When experimenting with pyramid architectures (eg: Swin), we need to explicitly preserve structure when downsampling, and how we do this is also described in Appendix.

\section{Experiments}
\label{sec:results}

In this section, we evaluate our family of models, SWAT on image classification and semantic segmentation. We use Imagenet-1K \cite{deng2009imagenet} and ADE20K \cite{zhou2019ade20k} as benchmarks to compare against common Transformer/Mixer/Conv architectures such as DeiT \cite{touvron2021deit}, Swin \cite{liu2021swin}, MLP-Mixer \cite{tolstikhin2021mixer}, ResMLP \cite{touvron2021resmlp} and VAN \cite{guo2022van}. In our ablations, we further evaluate the benefits of preserving structure.

\subsection{ImageNet Classification}
\label{subsec:charades}

ImageNet-1K \cite{deng2009imagenet} is a commonly-used classification benchmark, with 1.2M training images and 50K validation images, annotated with 1000 categories. For all our models, we report Top-1 (\%) accuracy on single-crop evaluation with complexity metrics such as Parameters and FLOPs. %
We train all our models for 300 epochs on inputs of $224\times224$ using the \texttt{timm} \cite{rw2019timm} library. We use the original hyperparameters for all backbones, without further tuning. 
All models are trained with Mixed Precision.

\begin{table}[t!]
	\centering
	\tablestyle{1.8pt}{1.}
	\fontsize{9}{11}\selectfont
		\begin{tabular}{llrrr}
            \toprule
			\multicolumn{1}{l}{\multirow{2}{*}{Model}}  & Model & Top-1 & Params. & FLOPs \\
			{} & scale & (\%) & (M) & (G) \\
			\toprule
			
			\multirow{3}{*}{DeiT (\citeauthor{touvron2021deit})} & Ti & 72.2 & 5.7 & 1.3 \\
             & S & 79.8 & 22.1 & 4.6 \\
             & B/32 & 75.5 & 88.2 & 4.3 \\
            
			\rowcolor{row} & Ti & \textbf{\scriptsize(+3.5)} 75.7 & 5.8 & 1.4 \\
			\rowcolor{row} & S & \textbf{\scriptsize(+0.7)} 80.5 & 22.3 & 4.9 \\
			\rowcolor{row}\multirow{-3}{*}{SWAT$_\text{DeiT}$ (ours)} & B/32 & \textbf{\scriptsize(+0.7)} 76.2 & 86.3 & 4.5 \\ \midrule
			
			\multirow{3}{*}{Mixer (\citeauthor{tolstikhin2021mixer})} & Ti & 68.3 & 5.1 & 1.0 \\
             & S & 75.7  & 18.5 & 3.8 \\
             & B/32 & 75.5 & 60.3 & 3.2 \\
            
			\rowcolor{row} & Ti & \textbf{\scriptsize(+4.0)} 72.3 & 5.1 & 1.0 \\
			\rowcolor{row} & S & \textbf{\scriptsize(+2.2)} 77.9 & 18.6 & 3.8 \\
 			\rowcolor{row} \multirow{-3}{*}{SWAT$_\text{Mixer}$ (ours)} & B/32 & \textbf{\scriptsize(+1.6)} 77.1 & 58.4 & 3.3 \\ \midrule

            \multirow{2}{*}{Swin (\citeauthor{liu2021swin})} & Ti & 81.3  & 28.3 & 4.5 \\
             & S & 83.0  & 49.6 & 8.7 \\
            
			\rowcolor{row} & Ti & \textbf{\scriptsize(+0.4)} 81.7 & 27.1 & 4.7 \\
			\rowcolor{row} \multirow{-2}{*}{SWAT$_\text{Swin}$ (ours)} & S & \textbf{\scriptsize(+0.3)} 83.3 & 48.9 & 9.1  \\
   \bottomrule

	\end{tabular}
	\caption{\textbf{SWAT is generally-applicable and scalable.} We compare SWAT with DeiT \protect\cite{touvron2021deit}, MLP-Mixer \protect\cite{tolstikhin2021mixer} and Swin \protect\cite{liu2021swin}  on ImageNet-1K. We report the performance in Tiny, Small and Base/32 (i.e., patch size of 32$\times$32) configurations. SWAT models consistently outperform their counterparts with minimal change in parameters or computations. %
	All models are trained for 300 epochs at $224\times224$ resolution. Performance improvement is in \textbf{bold}.
	}
	\vspace{-2mm}
	\label{tab:deit_mixer}
\end{table}

\begin{table}[ht!]
	\centering
	\tablestyle{1.8pt}{1.}
	\fontsize{8.8}{11}\selectfont
		\begin{tabular}{llrrr}
            \toprule
			\multicolumn{2}{l}{\multirow{2}{*}{Model}} & Top-1 & Params. & FLOPs \\
			 & & (\%) & (M) & (G) \\
			\toprule
			
			\multirow{6}{*}{\rotatebox[origin=c]{90}{CNN}} & ResNet (\citeauthor{he2016deep}) & 78.8 & 25.6 & 4.1 \\
            
            & ResNeXt* (\citeauthor{xie2017aggregated}) & 77.6 & 25.0 & 4.3 \\
            
            & EfficientNet* (\citeauthor{tan2019efficientnet}) & 82.6 & 19.3 & 4.4 \\
            
            & RegNetY* (\citeauthor{radosavovic2020designing}) & 79.4 & 20.6 & 4.0 \\
            
            & ConvMixer (\citeauthor{trockman2022patches}) & 80.2 & 21.1 & - \\
            
             &ConvNeXt (\citeauthor{liu2022convnet}) & 82.1 & 29.0 & 4.5 \\ \midrule
            
			\multirow{6}{*}{\rotatebox[origin=c]{90}{MLP}} & Mixer (\citeauthor{tolstikhin2021mixer}) & 75.7  & 18.5 & 3.8 \\
			\rowcolor{row}\cellcolor{white}& SWAT$_\text{Mixer}$ (ours) & \textbf{\scriptsize(+2.2)} 77.9 & 18.6 & 3.8 \\

            & gMLP (\citeauthor{touvron2021resmlp}) & 79.6 & 20.0 & 4.5 \\
			
			& ResMLP* (\citeauthor{touvron2021resmlp}) & 76.6 & 15.4 & 3.0 \\
			\rowcolor{row}\cellcolor{white} & SWAT$_\text{ResMLP}$* (ours) & \textbf{\scriptsize(+1.2)} \ch{77.8} & 15.6 & 3.1 \\
			
			& PoolFormer (\citeauthor{yu2022metaformer}) & 80.3 & 21.4 & 3.6 \\
			
			& CycleMLP (\citeauthor{chen2021cyclemlp}) & 81.6 & 27.0 & 3.9 \\ \midrule
            
			\multirow{10}{*}{\rotatebox[origin=c]{90}{Attention}} & DeiT (\citeauthor{touvron2021deit}) &  79.8 & 22.1 & 4.6 \\
			\rowcolor{row}\cellcolor{white} & SWAT$_\text{DeiT}$ (ours) & \textbf{\scriptsize(+0.7)} 80.5 & 22.3 & 4.9 \\

            & T2T-ViT (\mbox{\citeauthor{yuan2021tokens}}) & 81.5 & 21.5 & 4.8 \\
			
			& TNT (\citeauthor{han2021transformer}) & 81.5 & 23.8 & 5.2 \\
			
			& NesT (\citeauthor{zhang2021aggregating}) & 81.5 & 17.0 & 5.8 \\
			
			& PVT (\citeauthor{wang2021pyramid})  &79.8 & 24.5 & 3.8 \\
			
			& Twins (\citeauthor{chu2021twins}) & 81.7 & 24.0 & 2.8 \\
			
			& Focal (\citeauthor{yang2021focal}) & 82.2 & 29.1 & 4.9 \\
			
			& Swin (\citeauthor{liu2021swin}) & 81.3 & 28.3 & 4.5 \\
			\rowcolor{row}\cellcolor{white} & SWAT$_\text{Swin}$ (ours) & \textbf{\scriptsize(+0.4)} \ch{81.7} & 27.1 & 4.7 \\ \midrule

			\multirow{7}{*}{\rotatebox[origin=c]{90}{Hybrid}} & ConViT (\citeauthor{d2021convit}) &  81.3 & 27.0 & 5.4 \\
			
			& CvT (\citeauthor{wu2021cvt}) & 81.6 & 20.0 & 4.5 \\
			
			& Conformer (\citeauthor{peng2021conformer}) & 81.3 & 23.5 & 5.2 \\

			& CeiT (\citeauthor{yuan2021incorporating}) & 82.0 & 24.2 & 4.8 \\
			
			& MobileFormer* (\citeauthor{chen2022mobile}) & 79.3 & 14.0 & 0.5 \\
			
			& VAN (\citeauthor{guo2022van}) & 82.8 & 26.6 & 5.0 \\
			\rowcolor{row}\cellcolor{white} & SWAT$_\text{VAN}$ (ours) & \textbf{\scriptsize(+0.6)} \ch{83.4} & 27.0 & 5.8\\
            \bottomrule

	\end{tabular}
	\caption{\textbf{SWAT is competitive with SOTA.} We report experiments on ImageNet-1K with different families of token-based models in mid-sized configurations (14-30M params.). We implement SWAT with DeiT \protect\cite{touvron2021deit}, Swin \protect\cite{liu2021swin}, MLP-Mixer \protect\cite{tolstikhin2021mixer}, ResMLP \protect\cite{touvron2021resmlp} and VAN \protect\cite{guo2022van} baselines, and train with original hyperparameter settings. SWAT outperforms all baselines consistently with minimal change in complexity, showing competitive performance with SOTA models. In general, models are trained for 300 epochs at $224\times224$ resolution (exceptions denoted with $^*$ are discussed in appendix). Performance improvement is in \textbf{bold}.
	}
    \vspace{-3mm}
	\label{tab:midsize}
\end{table}

\paragraph{SWAT is generally-applicable and scalable:} In \tref{tab:deit_mixer}, we present the performance of SWAT with the two main types of token-based models: those using attention (MHSA) such as DeiT \cite{touvron2021deit} and Swin \cite{liu2021swin}, or those using MLPs such as Mixer \cite{tolstikhin2021mixer}. In both model families, SWAT consistently outperforms the baselines across different model scales, verifying that our Structure-aware Tokenization and Structure-aware Mixing can be applied in both cases.
Specifically, we consider Tiny, Small and Base/32 (i.e., patch size of 32$\times$32) model scales, with varying range of parameters and computations. These are standard models reported in previous work. We implement our tokenizer and replace Transformer/Mixing blocks with ours in each configuration (eg: DeiT-Ti $\rightarrow$ SWAT$_\text{DeiT}$-Ti). In all configurations, SWAT models show consistent improvements. In SWAT$_\text{DeiT}$, Tiny version achieves the highest gain of $+3.5\%$, with $+0.7\%$ in Small and Base/32. In SWAT$_\text{Mixer}$, all Tiny ($+4.0\%$), Small ($+2.2\%$) and Base/32 ($+1.6\%$) versions show a considerable improvement over baselines. SWAT$_\text{Swin}$ shows $+0.4\%$ w/ Tiny and $+0.3\%$ w/ Small models. Overall, SWAT models have minimal (or no) increment in parameters or computations. The performance vs. complexity graphs are shown in \fref{fig:acc_vs_compute}. 

\paragraph{SWAT is competitive with SOTA:} In \tref{tab:midsize}, we implement SWAT with multiple families of token-based models, either Transformer/Mixer/Convolutional, including DeiT \cite{touvron2021deit}, Swin \cite{liu2021swin}, Mixer \cite{tolstikhin2021mixer}, ResMLP \cite{touvron2021resmlp} and VAN \cite{guo2022van}. We report the performance in mid-sized (14-30M parameters) standard configurations. We use the same hyperparameter settings and training recipes as the corresponding original baselines.
We observe consistent gains in SWAT family of models: $+2.2\%$ in SWAT$_\text{Mixer}$ and $+1.2\%$ in SWAT$_\text{ResMLP}$, $+0.7\%$ in SWAT$_\text{DeiT}$, $+0.4\%$ in SWAT$_\text{Swin}$ and $+0.6\%$ in SWAT$_\text{VAN}$, 
with minimal change in parameters and computations compared to baselines. This further shows that SWAT can be generally-adopted to any token-based architecture with minimal effort and cost.

\begin{figure*}[t!]
	\centering
	\includegraphics[width=1\linewidth]{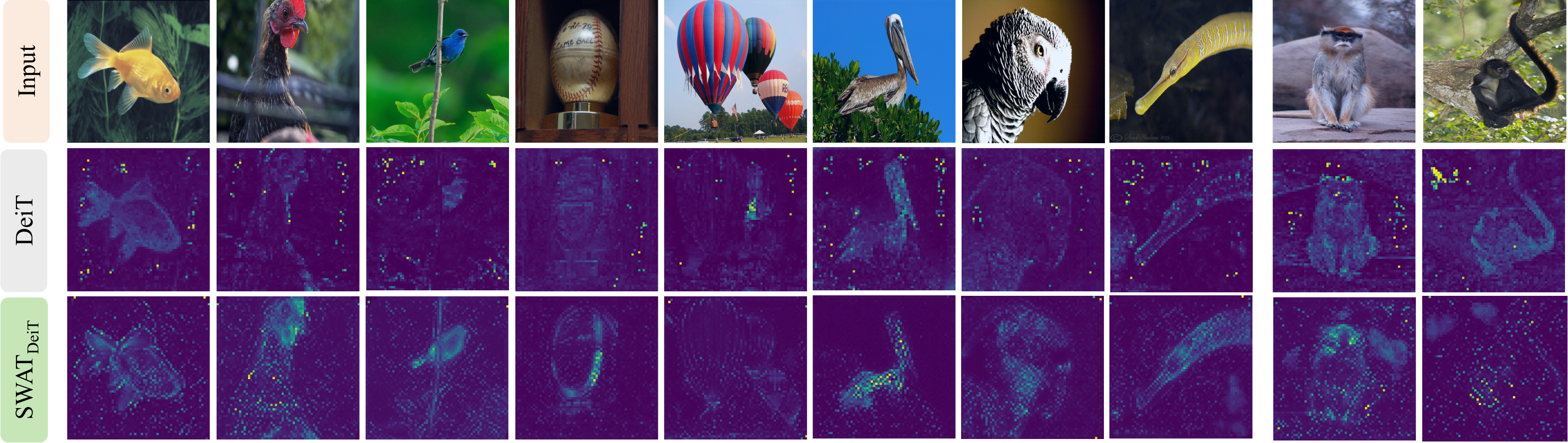}
	\caption{\textbf{Visualization of token attention} in DeiT-Ti \protect\cite{touvron2021deit} and SWAT$_\text{DeiT}$-Ti. We use the code from DINO \protect\cite{caron2021emerging} for visualization. However since we do not use class tokens as in DINO, we show the attention averaged across tokens. SWAT, as it preserves structure even within tokens, shows more contrastive and fine-grained attention maps compared to DeiT (even though we consider same number of tokens, i.e., resolution, in both). Note the better-visible boundaries and segments in SWAT, compared to smoothed-out variations in DeiT. Some cases where SWAT fails to capture fine details are also shown (to the right). Best viewed in color and zoomed-in.
	}
	\label{fig:attn_vis}
\end{figure*}

\paragraph{SWAT shows more fine-grained attention patterns:} In \fref{fig:attn_vis}, we visualize token attention values in Tiny configurations of DeiT \cite{touvron2021deit} and SWAT$_\text{DeiT}$. We use the code from DINO \cite{caron2021emerging} paper as a base. However, in our models, since we do not use a class token, we cannot visualize the attention on a single token as in \cite{caron2021emerging}. Instead, we show the attention maps of the final layer of each model, averaged across tokens. We consider larger image size ($1024\times1024$) compared to training ($224\times224$) to get higher resolution visualizations. We use the same patch size of 16 and interpolate positional encodings accordingly. We can see clear differences between the attention in DeiT \cite{touvron2021deit} and SWAT$_\text{DeiT}$. In SWAT, we have more contrastive attention which resembles fine-grained structures (eg: boundaries in object segments), since we preserve such structure within tokens. In contrast, DeiT attention is smoothed-out and subtle. Also, the attention weights in the SWAT model are less-noisy. We use the same resolution (i.e., same number of tokens) in both cases.

We include a detailed analysis of model throughput (im/s) in SWAT models and their baselines at inference, in the Appedix. We consider FLOPs and parameters as our metrics of complexity, as they are system-agnostic and reproducible.

\subsection{Ablations on ImageNet}

\begin{table}[t!]
	\centering
    \tablestyle{1.8pt}{1.}
	\fontsize{9}{11}\selectfont
		\begin{tabular}{lcccrrr}
            \toprule
    		\multicolumn{1}{l}{\multirow{2}{*}{Model}}  & \multicolumn{3}{c}{Structure-aware} & Top-1 & Params. &  FLOPs \\
            \cmidrule{2-4}
    		 & \multicolumn{1}{c}{Tokenize} & Tk. Mix. & \multicolumn{1}{c}{Ch. Mix.} & (\%) & (M) & (G) \\
    		\toprule
    		\multirow{1}{*}{DeiT} & \xmark & \xmark & \xmark & 73.3 & 5.72 & 1.25 \\ \midrule
    	     & \xmark & \xmark & \cmark & 74.6 & 5.96 & 1.30 \\
    	     & \xmark & \cmark & \xmark & 73.7 & 5.72 & 1.30 \\
    	     & \cmark & \xmark & \xmark & 73.0 & 5.58 & 1.30 \\
    	     & \cmark & \cmark & \xmark & 74.5 & 5.59 & 1.35 \\

    		 \rowcolor{row}SWAT$_\text{DeiT}$ & \cmark & \cmark & \cmark & \textbf{75.7}  & 5.83 & 1.40 \\
    		\bottomrule
    		\multirow{1}{*}{Mixer} & \xmark & \xmark & \xmark &  68.3 & 5.07  & 0.97 \\ \midrule
    		 & \xmark & \xmark & \cmark & 70.8 & 5.28 & 1.01 \\
    		 & \xmark & \cmark & \xmark & 68.9 & 5.08 & 0.97 \\
    	     & \cmark & \xmark & \xmark & 67.9 & 4.88 & 0.94 \\
    	     & \cmark & \cmark & \xmark & 70.2 & 4.88 & 0.95 \\
    		\rowcolor{row}SWAT$_\text{Mixer}$ & \cmark & \cmark & \cmark & \textbf{72.3} & 5.10 & 0.99 \\
            \bottomrule
    	\end{tabular}
    \vspace{-1mm}
	\caption{\textbf{Ablations on Structure with DeiT-Ti \protect\cite{touvron2021deit} and Mixer-Ti \protect\cite{tolstikhin2021mixer}} on ImageNet-1K. We report the gains from (1) Structure-aware Tokenization, (2) Token Mixing with Channel Structure, and (3) Channel Mixing with Token Structure. Structure-aware inputs and Structure-aware Mixing gives consistent improvements, as shown in \textbf{bold}. A key observation: our Tokenization and our Token Mixing should \textit{always} be coupled. %
	}
	\vspace{-3mm}
	\label{tab:ablation_deit}
\end{table}

In this section, We present ablations on Tiny versions of SWAT$_\text{DeiT}$ and SWAT$_\text{Mixer}$. Specifically, in \tref{tab:ablation_deit}, we focus on Structure-aware Tokenization, Token Mixing with Channel Structure and Channel Mixing with Token Structure. 

\paragraph{Structure-aware Tokenization:} We compare different settings with SWAT tokenizer. Bottom line is that \textit{Structure-aware Tokenization should always be coupled with the Structure-aware Token Mixing}. It makes sense: if we prepare tokens with structure and not take advantage of it during mixing, it does not really have a benefit and the reduced capacity (due to our Tokenization) may even drop the performance.
In DeiT \cite{touvron2021deit}, we see such performance drop of $-0.3\%$ when we do not use the structure (\textit{within} tokens) explicitly.
In Mixer \cite{tolstikhin2021mixer}, this drop is $-0.4\%$. In both cases, when we specifically make use of the newly-introduced structure \textit{within} tokens, we see consistent gains ($+1.5\%$ in DeiT and $+2.3\%$ in Mixer) over our Tokenization-only versions. 

\paragraph{Channel Structure (\textit{within} tokens) in Token Mixing:} Here, we intend to consider a local neighborhood within tokens. Even if such a structure is not present (i.e., not having our Tokenization), models can benefit slightly: $+0.4\%$ in DeiT \cite{touvron2021deit} and $+0.6\%$ in Mixer \cite{tolstikhin2021mixer}. This is due to the inductive bias of replacing Linear layers with Conv. However, the true potential of this comes when a structure \textit{within} tokens is explicitly available, where we see a $+1.2\%$ improvement in DeiT \cite{touvron2021deit} and $+1.9\%$ in Mixer \cite{tolstikhin2021mixer}.

\paragraph{Token Structure (\textit{among} tokens) in Channel Mixing:} Here, we consider a local neighborhood among tokens. In DeiT \cite{touvron2021deit}, we see $+1.3\%$ boost, and in Mixer \cite{tolstikhin2021mixer}, a $+2.5\%$ boost in performance.

\subsection{Semantic Segmentation}
\label{subsec:multithumos}

ADE20K \cite{zhou2019ade20k} benchmark contains annotations for semantic segmentation across 150 categories. It comes with 25K annotated images in total, with 20K training, 2K validation and 3K testing. We report mIoU for our models in multi-scale testing (i.e., [0.5, 0.75, 1.0, 1.25, 1.5, 1.75]$\times$ the training resolution) similar to previous work \cite{liu2021swin}, along with complexity metrics such as parameters, FLOPs (for input size of $512\times2048$ similar to \cite{liu2021swin}) and frame-rate.
We follow a similar training recipe to Swin \cite{liu2021swin}. Our backbones are pretrained on ImageNet-1K \cite{deng2009imagenet} for 300 epochs at $224\times224$, before re-training with a decoder for segmentation at $512\times512$. We use UperNet \cite{xiao2018unified} as our decoder within \texttt{mmsegmentation} \cite{mmseg2020} framework. We use the original hyperparameter settings as the baseline.

\begin{table}[t!]
	\centering
	\tablestyle{1.8pt}{1.}
	\fontsize{8.8}{11}\selectfont
	\resizebox{1.\linewidth}{!}{
		\begin{tabular}{llrrrr}
            \toprule
			\multicolumn{1}{l}{\multirow{2}{*}{Method}}  & \multicolumn{1}{l}{\multirow{2}{*}{Backbone}} & \multicolumn{1}{c}{\multirow{2}{*}{mIoU}} & Params. & FLOPs & \multirow{2}{*}{FPS} \\
			& & & (M) & (G) & \\
			\toprule
			DANet (\citeauthor{fu2019dan}) & \multirow{6}{*}{Resnet-101 (\citeauthor{he2016deep})} & 45.2 & 69 & 1119 & 15.2 \\
			DLab.v3+ (\citeauthor{chen2018encoder}) & & 44.1 & 63 & 1021 & 16.0 \\
			ACNet (\citeauthor{fu2019adaptive}) & & 45.9 & - & - & - \\
			DNL (\citeauthor{yin2020disentangled}) & & 46.0 & 69 & 1249 & 14.8 \\
			OCRNet (Yuan et al.) & & 45.3 & 56 & 923 & 19.3 \\
			UperNet (\citeauthor{xiao2018unified}) & & 44.9 & 86 & 1029 & 20.1 \\ \midrule
			 & DeiT-S (\citeauthor{touvron2021deit}) & 44.0 & 52 & 1099 & 16.2 \\
			 & Swin-Ti (\citeauthor{liu2021swin}) & 45.8 & 60 & 945 & 18.5 \\
			 \rowcolor{row}\cellcolor{white} & SWAT$_\text{Swin}$-Ti (ours) & \textbf{46.5} & 59 & 950 & 16.9 \\

             & VAN-B2 (\citeauthor{guo2022van}) & 50.1 & 57 & 948 & - \\
			\rowcolor{row}\cellcolor{white} \multirow{-5}{*}{UperNet (\citeauthor{xiao2018unified})} & SWAT$_\text{VAN}$-B2 (ours) & \textbf{50.7} & 55 & 952 & - \\
            \bottomrule
			
	\end{tabular}
	}
	\vspace{-1mm}
	\caption{\textbf{SWAT for semantic segmentation} on ADE20K \protect\cite{zhou2019ade20k} dataset. We report results in the same setting as Swin \protect\cite{liu2021swin} using \texttt{mmsegmentation} \protect\cite{mmseg2020} framework. 
	FPS is measured on a single V100 GPU. SWAT outperforms respective baselines, but is slightly slower due to extra convolutions.}
	\vspace{-2mm}
	\label{tab:seg}
\end{table}

\paragraph{Results:} In \tref{tab:seg}, we show the performance of SWAT$_\text{Swin}$ and SWAT$_\text{VAN}$ backbones when used with the UperNet \cite{xiao2018unified} head for semantic segmentation on ADE20K, and compare with similar-sized baselines. 
SWAT$_\text{Swin}$ gives $+0.7$ mIoU and SWAT$_\text{VAN}$ gives $+0.6$ mIoU improvement over the respective baselines, when trained under the same settings. However, FPS of the SWAT$_\text{Swin}$ based model is slightly lower, due to extra convolutions introduced in SWAT. %
In the Appendix, we include segmentation masks generated by SWAT$_\text{Swin}$ and Swin \cite{liu2021swin} backbones, which qualitatively show this improvement.

\section{Conclusion}
\label{sec:conclusion}

In this work, we present the merits of preserving spatial structure, both \textit{within} and \textit{among} Tokens, in common Transformer/Mixer/Convolutional token-based architectures. Our two key contributions are: (1) \textit{Structure-aware Tokenization} and (2) \textit{Structure-aware Mixing}, which can be adopted in different families of models with minimal effort. The resulting family of models, SWAT, outperforms the corresponding baselines and shows competitive performance with SOTA models on multiple benchmarks, with minimal change in parameters and computations. We hope that SWAT will open-up new ways of making use of spatial structure as an inductive bias in token-based models. %

\section*{Acknowledgements}
This work was supported by the National Science Foundation (IIS-2104404 and CNS-2104416). We thank the Robotics Lab at SBU for helpful discussions.

\section{Appendix}
\label{sec:appendix}

\setcounter{table}{0}
\setcounter{figure}{0}
\renewcommand{\thetable}{A.\arabic{table}}
\renewcommand{\thefigure}{A.\arabic{figure}}

\subsection{Discussion and Ablations}

\paragraph{Why parallel 2D Conv in Transformer Token Mixing?} Another option is to replace the Linear layers in Transformer token mixing block with 2D Conv directly, without the parallel design. However, if we do this, the number of parameters reduce considerably (as $c$ is small), impairing the model capacity. Thus, we have a parallel design (with a negligible parameter increase due the Conv layer), and sum the outputs, propagating the channel structure through the newly-added branch. Here, outputs of each branch is scaled by $\times0.5$ to avoid any training instability in MHSA.  

\paragraph{Pyramid architectures:} It is rather straightforward to implement SWAT tokenization in homogeneous structures (w/ uniform resolution) such as ViT \cite{dosovitskiy2020image}, DeiT \cite{touvron2021deit} or Mixers \cite{tolstikhin2021mixer}. However, we also experiment with pyramid structures (w/ progressive-downsampling) such as Swin \cite{liu2021swin}. Here, SWAT tokenization is applied at the input-level as before, without any change. It is just that we also need to preserve the structure (both within and among tokens) while downsampling. To do this, we implement a new patch-merging operation. In the Swin \cite{liu2021swin} baseline, patch-merging maps
$B \times H\times W\times C \rightarrow B \times H/2\times W/2\times 4C \rightarrow B \times H/2\times W/2\times 2C$ 
with reshaping and Linear layers, applied sequentially. It may break our structure \textit{within} tokens (as no skip connections either to preserve it). Therefore, instead, we first reshape as $B \times H\times W\times C \rightarrow B \times H\times W\times (chw) \rightarrow B \times c\times (Hh)\times (Ww)$ to the original 2D structure, and then apply a strided 2D Conv with a stride of 2. Finally, we reshape the structure back into the channels as $B \times 2c\times (Hh/2)\times (Ww/2) \rightarrow B \times H/2\times W/2\times 2(chw)$.

\paragraph{On the spatial structure \textit{within} tokens:} We consider different settings for the \ch{structure hyperparameter $\alpha$} (ablated in \tref{tab:ablation_alpha_kernel}), based on the embedding dimension $C$ and the patch size $p$ of the architecture. \ch{It decides the the number of sub-tokens ($\alpha^2$) preserved within a token.}. As default settings, we consider a structure of $8\times8$ (i.e., $\alpha=8$) for models with a patch size of $16\times16$, (eg: DeiT Tiny and Small), a structure of $16\times16$ for patch size of $32\times32$ (eg: DeiT Base/32), and a structure of $2\times2$ for patch size of $4\times4$ (eg: Swin Tiny). We set the channel size of the intermediate tokens to fit the embedding dimension of the model.

\ch{\paragraph{Granularity of preserved structure:} We use the structure hyperparameter $\alpha$ to control the granularity of preserved structure within tokens. In fact, we preserve $\alpha\times\alpha$ sub-tokens as channel segments in our Tokenization. In \tref{tab:ablation_alpha_kernel}, we consider $\alpha=\{8,4,2\}$ and it shows that the finest structure within tokens gives the best performance.}

\begin{figure*}[t]
	\centering
	\includegraphics[width=1\linewidth]{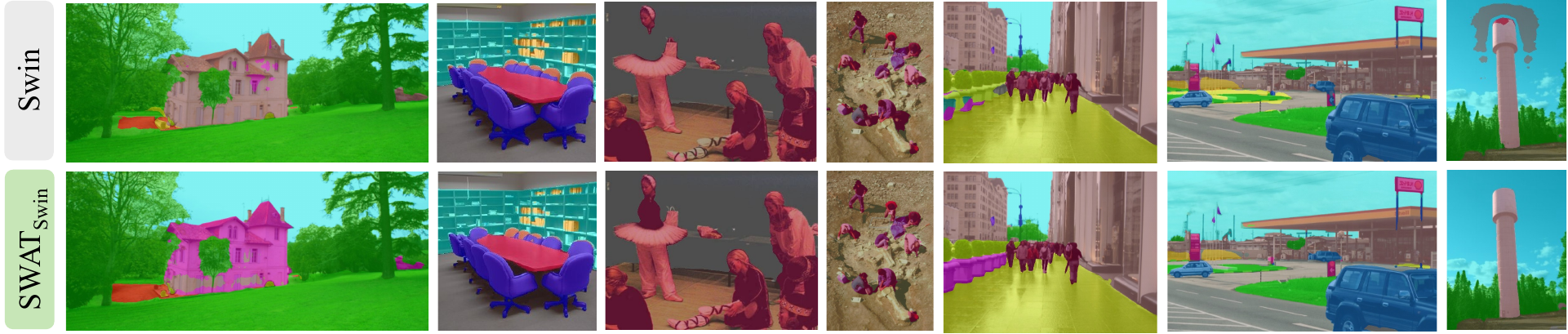}
	\caption{\textbf{Segmentation masks} generated with UperNet \protect\cite{xiao2018unified} with Swin-Ti \protect\cite{liu2021swin} (our setting) and SWAT$_\text{Swin}$-Ti as backbones. SWAT shows better segmentation results in comparison. Note the fine structures better captured by SWAT, thanks to preserved spatial information within tokens. Best viewed in color and zoomed-in.}
	\vspace{-2mm}
	\label{fig:seg}
\end{figure*}

\begin{table}[ht!]
	\centering
    \tablestyle{1.8pt}{1.}
	\fontsize{9}{11}\selectfont
		\begin{tabular}{lccrr}
            \toprule
    		\multicolumn{1}{l}{Model}  & $\;\;\alpha\;\;$ & Kernel & Top-1 &  FLOPs \\
            \toprule
    		Mixer (\citeauthor{tolstikhin2021mixer}) & - & - &  68.3\%  & 0.97G \\ %
            \midrule
    		& 2 & $(5\times5)$ & 70.9\% & 0.98G \\ 
    		& 4 & $(5\times5)$ & 71.1\% & 0.99G \\
    		\rowcolor{row} \cellcolor{white}\multirow{1}{*}{SWAT$_\text{Mixer}$}& 8 & $(5\times5)$ & \textbf{72.3}\% & 0.99G \\
    		& 8 & $(3\times3)$ & 69.8\% & 0.96G \\
    		& 8 & $(7\times7)$ & 71.2\% & 1.02G \\
            \bottomrule
    	\end{tabular}
    \vspace{-1mm}
	\caption{\textbf{Ablations on (1) the granularity of preserved structure, and (2) the level of structure-awareness in mixing}, with SWAT$_\text{Mixer}$-Ti models on ImageNet-1K. With structure hyperparameter $\alpha$, we preserve an $\alpha\times \alpha$ structure within tokens. Increasing granularity (i.e., preserving finer details) shows consistent performance gains. Kernel size in Mixing decides how-well we utilize the structure (i.e., structure-awareness). In Token Mixing, we fix Conv kernel to be $(3\times3)$ due to spatial size constraints. In Channel mixing, we consider kernel sizes of \{3,5,7\}. Conv kernels of $(5\times5)$ shows the best performance at a similar complexity as the baseline.
	}
	\vspace{-2mm}
	\label{tab:ablation_alpha_kernel}
\end{table}

\ch{\paragraph{Receptive field when utilizing structure:} We always consider $3\times3$ Conv kernels for token mixing (since the input spatial dimension $h\times w$ within tokens is only $8\times8$) in these settings. However, we can consider kernel sizes of $\{3,5,7\}$ in channel mixing (since the input spatial dimension $H/p\times W/p$ among tokens is $14\times14$). Higher kernel size means we exploit more structure in Mixing, but also, it increases both parameters and computations as well. In \tref{tab:ablation_alpha_kernel}, we see that $5\times5$ kernels shows the best performance at a similar complexity as in the baseline Mixer \cite{tolstikhin2021mixer}.}

\ch{\paragraph{SWAT regular patch-size vs. Baseline small patch-size:} If we simply consider smaller patch-sizes to preserve fine-grained information instead of a SWAT-like implementation, it will either blow-up the computations or lack the model capacity. For instance, a transformer block has $12C^2$ parameters, whereas the FLOP count scales with $(12NC^2 + 2N^2C)$. Here, $C$ is number of channels and $N$ is number of tokens. If we directly reduce the patch-size, it will give more tokens, increasing the computations quadratically. We can contain this by reducing the channel size, but it will also reduce the number of parameters, sacrificing the model capacity. In contrast, SWAT retains the fine-grained information similar to a model with smaller patch-size, but at the same footprint as a model with original patch-size.}

\paragraph{Segmentation masks:} \fref{fig:seg} shows qualitative results of SWAT$_{\text{Swin}}$ on semantic segmentation, compared to its baseline. Our model achieves better segmentation masks with a focus on fine-grained details and coherent structure. 

\paragraph{Exceptions to the default training schedule and input resolution:} In general, we compare against models trained for 300 epochs at $224\times224$ resolution. However, some baselines in Table 2 have different settings, namely, ResNeXt: 90 epochs, RegNetY: 100 epochs, ResMLP: 400 epochs, MobileFormer: 450 epochs and EfficientNet: $380\times380$ resolution.

\subsection{Throughput of SWAT models}

\begin{table}[t!]
	\centering
	\tablestyle{1.8pt}{1.}
	\fontsize{8.8}{11}\selectfont
	\begin{tabular}{lrrrr}
            \toprule
			\multicolumn{1}{l}{\multirow{2}{*}{Model}} & Top-1 & Params. & FLOPs & Throughput \\
			 & (\%) & (M) & (G) & (im/s) \\
			\toprule
			
			DeiT - Ti (\citeauthor{touvron2021deit}) & 72.2 & 5.7 & 1.3 & 2391  \\
			\rowcolor{row}SWAT$_\text{DeiT}$ - Ti & \textbf{75.7} & 5.8 & 1.4 & 1330 \\ %
			\midrule
			
			DeiT - S (\citeauthor{touvron2021deit}) & 79.8 & 22.1 & 4.6 & 924 \\
			\rowcolor{row}SWAT$_\text{DeiT}$ - S & \textbf{80.5} & 22.3 & 4.9 & 645 \\ %
			\midrule
			
			DeiT/32 - B (\citeauthor{touvron2021deit}) & 75.5 & 88.2 & 4.3 & 1275 \\
			\rowcolor{row}SWAT$_\text{DeiT}$/32 - B & \textbf{76.2} & 86.3 & 4.5 &  923 \\ %
			\midrule
			
			Mixer - Ti (\citeauthor{tolstikhin2021mixer}) & 68.3 & 5.1 & 1.0 & 3701 \\
			\rowcolor{row}SWAT$_\text{Mixer}$ - Ti & \textbf{72.3} & 5.1 & 1.0 & 2759 \\  %
			\midrule
			
			Mixer - S (\citeauthor{tolstikhin2021mixer}) & 75.7  & 18.5 & 3.8 & 1235  \\
			\rowcolor{row}SWAT$_\text{Mixer}$ - S & \textbf{77.9} & 18.6 & 3.8 & 1022 \\ %
			
			\midrule
			
			Swin - Ti (\citeauthor{liu2021swin}) & 81.3 & 28.3 & 4.5 & 703  \\
			\rowcolor{row}SWAT$_\text{Swin}$ - Ti & \textbf{81.7} & 27.1 & 4.7 & 402 \\ %
			
			\midrule
			
			ResMLP (\citeauthor{touvron2021resmlp}) & 76.6 & 15.4 & 3.0 & 1562  \\
			\rowcolor{row}SWAT$_\text{ResMLP}$ & \textbf{77.8} & 15.6 & 3.1 & 1136 \\ %
            \bottomrule

	        \end{tabular}
	\vspace{-1mm}
	\caption{\textbf{Throughput analysis of different SWAT models:} We report performance and complexity (params., FLOPs and Throughput). System-agnostic metrics (params. and FLOPs) show consistent changes. However, throughput (on a single V100) show interesting variations: (1) smaller changes in larger models, and (2) changes in Transformer models are relatively higher than in Mixer models.}
	\vspace{-2mm}
	\label{tab:scale}
\end{table}

\begin{table}[t!]
	\centering
	\tablestyle{1.8pt}{1.}
	\fontsize{8.8}{11}\selectfont
	\begin{tabular}{lcccrrrr}
            \toprule
    		\multicolumn{1}{l}{\multirow{2}{*}{Model}}  & \multicolumn{3}{c}{Structure-aware} & Top-1 & Params. &  FLOPs & Thrput.\\
            \cmidrule{2-4}
    		 & \multicolumn{1}{c}{$\;$Tk.$\;$} & $\;$TM$\;$ & \multicolumn{1}{c}{$\;$CM$\;$} & (\%) & (M) & (G) & (im/s) \\
    		\toprule
    		\multirow{1}{*}{DeiT} & \xmark & \xmark & \xmark & 73.3 & 5.72 & 1.25 & 2391 \\
    		\midrule
    	     & \xmark & \xmark & \cmark & 74.6 & 5.96 & 1.30 &  1911 \\%
    	     & \xmark & \cmark & \xmark & 73.7 & 5.72 & 1.30 & 1778 \\ %
    	     & \cmark & \xmark & \xmark & 73.0 & 5.58 & 1.30 & 2008 \\ %
    	     & \cmark & \cmark & \xmark & 74.5 & 5.59 & 1.35 & 1551 \\ %

    		 \rowcolor{row}SWAT$_\text{DeiT}$ & \cmark & \cmark & \cmark & \textbf{75.7}  & 5.83 & 1.40 & 1330 \\
    		\bottomrule
    		\multirow{1}{*}{Mixer} & \xmark & \xmark & \xmark &  68.3 & 5.07  & 0.97 & 3701\\
    		\midrule
    		 & \xmark & \xmark & \cmark & 70.8 & 5.28 & 1.01 & 3053\\%
    		 & \xmark & \cmark & \xmark & 68.9 & 5.08 & 0.97 & 3534 \\ %
    	     & \cmark & \xmark & \xmark & 67.9 & 4.88 & 0.94 & 3452 \\%
    	     & \cmark & \cmark & \xmark & 70.2 & 4.88 & 0.95 & 3215 \\ %
    		\rowcolor{row}SWAT$_\text{Mixer}$ & \cmark & \cmark & \cmark & \textbf{72.3} & 5.10 & 0.99 & 2759 \\
      \bottomrule
    	\end{tabular}
	\vspace{-1mm}
	\caption{\textbf{Throughput analysis of SWAT ablations:} We report performance and complexity (params., FLOPs and Throughput). Each SWAT model shows small changes in system-agnostic metrics compared to the corresponding baseline. However, throughput changes are inconsistent in general. Our tokenization costs $-16\%$ in DeiT \protect\cite{touvron2021deit}, but only $-7\%$ in Mixer \protect\cite{tolstikhin2021mixer}.
	Our token mixing costs $-25\%$ throughput in DeiT, but only $-5\%$ in Mixer. In contrast, channel mixing costs are comparable, with $-20\%$ and $-17\%$, in DeiT and Mixer respectively.}
	\vspace{-2mm}
	\label{tab:ablation}
\end{table}

\begin{table}[t!]
	\centering
	\tablestyle{1.8pt}{1.}
	\fontsize{8.8}{11}\selectfont
	\begin{tabular}{lcrr}
		\toprule
		\multicolumn{1}{l}{\multirow{2}{*}{Mixer Block configuration}} & \multirow{1.5}{*}{Struct.} & \multicolumn{2}{c}{Throughput (im/s)} \\
		\cmidrule{3-4}
		& \multirow{-1.5}{*}{tokens} & \textbf{V100}  & \textbf{A100}\\
		\toprule
		$2\times$ Linear & \xmark & 3701 & 9127 \\
		$2\times$ 1D Conv (P) & \xmark & 3441 & 6921 \\
		$2\times$ 2D Conv (P) & \xmark & 3716 & 7421 \\
		\midrule
		$2\times$ 1D Conv (P) + $1\times$ Linear & \xmark & 2496 & 5480 \\
		$2\times$ 2D Conv (P) + $1\times$ Linear & \xmark & 2440 & 5290 \\
		$2\times$ 2D Conv (P) + $1\times$ 2D Conv (D) & \xmark & 2927 & 5538 \\
		\rowcolor{row}$2\times$ 2D Conv (P) + $1\times$ 2D Conv (D) & \cmark & 2759 & 5112 \\
		\bottomrule
		
	\end{tabular}
	\caption{\textbf{Throughput analysis of different SWAT$_\text{Mixer}$ block configurations:} We report performance and complexity (params., FLOPs and Throughput--- measured on a single V100/A100). Here, (P) represents Pointwise and (D), Depthwise Conv. Throughputs depend on the hardware, and different cuda implementations. On A100s, Linear layers are significantly faster than Conv (P). Even though Linear, 1D Conv (P) and 2D Conv (P) perform the same operation, they have different throughputs. Additional Linear layers incur higher cost compared to 2D Conv (D) on V100, which is not the case on A100.}
	\label{tab:block}
\end{table}

In this paper, we consider FLOPs as the main complexity metric, since it is system-agnostic. Here, we report throughput numbers using PyTorch 1.7.1 on a single V100 GPU. These may change depending on the actual hardware and underlying cuda optimizations.

\paragraph{With model scale:} When we consider larger SWAT models, the change of throughput due to Structure-aware Tokenization and Mixing becomes small (see \tref{tab:scale}). In DeiT \cite{touvron2021deit}, we see a $-44\%$ change in Tiny, $-30\%$ in Small and $-28\%$ in Base/32 models. We see a $-25\%$ change in Tiny and $-17\%$ in Small models in Mixer \cite{tolstikhin2021mixer}. However, none of the SWAT models have significant change in system-agnostic measures such as parameters or FLOPs.

\paragraph{In Transformers vs. Mixers:} The change in throughput with Transformer models such as DeiT-Ti ($-44\%$) \cite{touvron2021deit} or Swin-Ti \cite{liu2021swin} ($-43\%$) is higher compared to Mixer models such as Mixer-Ti \cite{tolstikhin2021mixer} ($-25\%$) or ResMLP-S12 \cite{touvron2021resmlp} ($-27\%$) (see \tref{tab:scale} and \tref{tab:ablation}). The difference between models is in Token Mixing. In Transformers, SWAT includes additional $2\times$ parallel 2D Conv blocks, whereas in Mixers, $1\times$ cascaded 2D Depthwise Conv block. 2D Conv blocks in Transformers see small feature depths (eg: depth$=3$ in Tiny) and run in parallel to Linear layers, which makes it hard to justify the extra throughput cost compared to cascaded 2D Depthwise Conv in Mixers. This shows inconsistency in cuda optimizations for different operators.

\paragraph{Using Linear vs. Conv:} Same operation as a Linear layer can be performed using 1D or 2D Pointwise Conv. However, in terms of throughput, Pointwise Conv layers show differences compared to Linear layers both on V100s ($-7\%$ w/ 1D and $0\%$ w/ 2D) and on A100s ($-24\%$ w/ 1D and $-19\%$ w/ 2D). See \tref{tab:block}. This shows differences in cuda optimizations, even for essentially-the-same operation.

\paragraph{On V100 vs. A100:} We see striking differences in throughput variations on V100 compared to A100 (see \tref{tab:block}). On A100s, implementation of Linear layers seems to be much faster compared to Pointwise Conv. This difference in throughput directly propagates to SWAT models.

Looking at the above inconsistencies in measuring complexity as throughput, we consider system-agnostic measures such as parameters or FLOPs to be more reliable and reproducible metrics to be evaluated in this paper.

\bibliographystyle{named}
\small
\bibliography{egbib}

\begin{thebibliography}{}

\bibitem[\protect\citeauthoryear{Arnab \bgroup \em et al.\egroup
  }{2021}]{arnab2021vivit}
Anurag Arnab, Mostafa Dehghani, Georg Heigold, Chen Sun, Mario Lu\v{c}i\'c, and
  Cordelia Schmid.
\newblock Vi{V}i{T}: {A} {V}ideo {V}ision {T}ransformer.
\newblock In {\em ICCV}, October 2021.

\bibitem[\protect\citeauthoryear{Bello}{2020}]{bello2021lambdanetworks}
Irwan Bello.
\newblock Lambda{N}etworks: {M}odeling long-range {I}nteractions without
  {A}ttention.
\newblock In {\em ICLR}, 2020.

\bibitem[\protect\citeauthoryear{Cao \bgroup \em et al.\egroup
  }{2021}]{cao2021video}
Jiezhang Cao, Yawei Li, Kai Zhang, and Luc Van~Gool.
\newblock Video {S}uper-{R}esolution {T}ransformer.
\newblock {\em arXiv:2106.06847}, 2021.

\bibitem[\protect\citeauthoryear{Carion \bgroup \em et al.\egroup
  }{2020}]{carion2020end}
Nicolas Carion, Francisco Massa, Gabriel Synnaeve, Nicolas Usunier, Alexander
  Kirillov, and Sergey Zagoruyko.
\newblock End-to-{E}nd {O}bject {D}etection with {T}ransformers.
\newblock In {\em ECCV}. Springer, 2020.

\bibitem[\protect\citeauthoryear{Caron \bgroup \em et al.\egroup
  }{2021}]{caron2021emerging}
Mathilde Caron, Hugo Touvron, Ishan Misra, Herv\'e J\'egou, Julien Mairal,
  Piotr Bojanowski, and Armand Joulin.
\newblock Emerging {P}roperties in {S}elf-{S}upervised {V}ision {T}ransformers.
\newblock In {\em ICCV}, October 2021.

\bibitem[\protect\citeauthoryear{Chen \bgroup \em et al.\egroup
  }{2018}]{chen2018encoder}
Liang-Chieh Chen, Yukun Zhu, George Papandreou, Florian Schroff, and Hartwig
  Adam.
\newblock Encoder-{D}ecoder with {A}trous {S}eparable {C}onvolution for
  {S}emantic {I}mage {S}egmentation.
\newblock In {\em ECCV}, 2018.

\bibitem[\protect\citeauthoryear{Chen \bgroup \em et al.\egroup
  }{2021a}]{chen2021decision}
Lili Chen, Kevin Lu, Aravind Rajeswaran, Kimin Lee, Aditya Grover, Michael
  Laskin, Pieter Abbeel, Aravind Srinivas, and Igor Mordatch.
\newblock Decision {T}ransformer: {R}einforcement {L}earning via {S}equence
  {M}odeling.
\newblock {\em URL Workshop in ICML}, 2021.

\bibitem[\protect\citeauthoryear{Chen \bgroup \em et al.\egroup
  }{2021b}]{chen2021cyclemlp}
Shoufa Chen, Enze Xie, Chongjian Ge, Ding Liang, and Ping Luo.
\newblock {CycleMLP: A MLP-like Architecture for Dense Prediction}.
\newblock {\em arXiv:2107.10224}, 2021.

\bibitem[\protect\citeauthoryear{Chen \bgroup \em et al.\egroup
  }{2022}]{chen2022mobile}
Yinpeng Chen, Xiyang Dai, Dongdong Chen, Mengchen Liu, Xiaoyi Dong, Lu~Yuan,
  and Zicheng Liu.
\newblock Mobile-former: Bridging mobilenet and transformer.
\newblock In {\em CVPR}, 2022.

\bibitem[\protect\citeauthoryear{Chu \bgroup \em et al.\egroup
  }{2021}]{chu2021twins}
Xiangxiang Chu, Zhi Tian, Yuqing Wang, Bo~Zhang, Haibing Ren, Xiaolin Wei,
  Huaxia Xia, and Chunhua Shen.
\newblock Twins: Revisiting the design of spatial attention in vision
  transformers.
\newblock {\em NeurIPS}, 34, 2021.

\bibitem[\protect\citeauthoryear{Dai \bgroup \em et al.\egroup
  }{2021}]{dai2021up}
Zhigang Dai, Bolun Cai, Yugeng Lin, and Junying Chen.
\newblock {UP-DETR}: {U}nsupervised {P}re-training for {O}bject {D}etection
  with {T}ransformers.
\newblock In {\em CVPR}, 2021.

\bibitem[\protect\citeauthoryear{Dai \bgroup \em et al.\egroup
  }{2022}]{dai2022ms}
Rui Dai, Srijan Das, Kumara Kahatapitiya, Michael~S Ryoo, and Francois Bremond.
\newblock {MS-TCT: Multi-Scale Temporal ConvTransformer for Action Detection}.
\newblock In {\em CVPR}, 2022.

\bibitem[\protect\citeauthoryear{Dauphin \bgroup \em et al.\egroup
  }{2017}]{dauphin2017language}
Yann~N Dauphin, Angela Fan, Michael Auli, and David Grangier.
\newblock Language {M}odeling with {G}ated {C}onvolutional {N}etworks.
\newblock In {\em ICML}. PMLR, 2017.

\bibitem[\protect\citeauthoryear{Deng \bgroup \em et al.\egroup
  }{2009}]{deng2009imagenet}
Jia Deng, Wei Dong, Richard Socher, Li-Jia Li, Kai Li, and Li~Fei-Fei.
\newblock Imagenet: A large-scale hierarchical image database.
\newblock In {\em CVPR}. IEEE, 2009.

\bibitem[\protect\citeauthoryear{Devlin \bgroup \em et al.\egroup
  }{2019}]{devlin2018bert}
Jacob Devlin, Ming-Wei Chang, Kenton Lee, and Kristina Toutanova.
\newblock {BERT}: Pre-training of deep bidirectional transformers for language
  understanding.
\newblock In {\em NAACL-HLT}. Association for Computational Linguistics, June
  2019.

\bibitem[\protect\citeauthoryear{Dosovitskiy \bgroup \em et al.\egroup
  }{2021}]{dosovitskiy2020image}
Alexey Dosovitskiy, Lucas Beyer, Alexander Kolesnikov, Dirk Weissenborn,
  Xiaohua Zhai, Thomas Unterthiner, Mostafa Dehghani, Matthias Minderer, Georg
  Heigold, Sylvain Gelly, et~al.
\newblock An {I}mage is {W}orth 16x16 {W}ords: {T}ransformers for {I}mage
  {R}ecognition at {S}cale.
\newblock {\em ICLR}, 2021.

\bibitem[\protect\citeauthoryear{Duke \bgroup \em et al.\egroup
  }{2021}]{duke2021sstvos}
Brendan Duke, Abdalla Ahmed, Christian Wolf, Parham Aarabi, and Graham~W
  Taylor.
\newblock {SSTVOS}: {S}parse {S}patiotemporal {T}ransformers for {V}ideo
  {O}bject {S}egmentation.
\newblock In {\em CVPR}, 2021.

\bibitem[\protect\citeauthoryear{d’Ascoli \bgroup \em et al.\egroup
  }{2021}]{d2021convit}
St{\'e}phane d’Ascoli, Hugo Touvron, Matthew~L Leavitt, Ari~S Morcos, Giulio
  Biroli, and Levent Sagun.
\newblock {ConViT: Improving Vision Transformers with Soft Convolutional
  Inductive Biases}.
\newblock In {\em ICML}. PMLR, 2021.

\bibitem[\protect\citeauthoryear{Esser \bgroup \em et al.\egroup
  }{2021}]{esser2021taming}
Patrick Esser, Robin Rombach, and Bjorn Ommer.
\newblock Taming {T}ransformers for {H}igh-{R}esolution {I}mage {S}ynthesis.
\newblock In {\em CVPR}, 2021.

\bibitem[\protect\citeauthoryear{Fan \bgroup \em et al.\egroup
  }{2021}]{fan2021multiscale}
Haoqi Fan, Bo~Xiong, Karttikeya Mangalam, Yanghao Li, Zhicheng Yan, Jitendra
  Malik, and Christoph Feichtenhofer.
\newblock Multiscale {V}ision {T}ransformers.
\newblock In {\em ICCV}, October 2021.

\bibitem[\protect\citeauthoryear{Fu \bgroup \em et al.\egroup
  }{2019a}]{fu2019dan}
Jun Fu, Jing Liu, Haijie Tian, Yong Li, Yongjun Bao, Zhiwei Fang, and Hanqing
  Lu.
\newblock Dual {A}ttention {N}etwork for {S}cene {S}egmentation.
\newblock In {\em CVPR}, 2019.

\bibitem[\protect\citeauthoryear{Fu \bgroup \em et al.\egroup
  }{2019b}]{fu2019adaptive}
Jun Fu, Jing Liu, Yuhang Wang, Yong Li, Yongjun Bao, Jinhui Tang, and Hanqing
  Lu.
\newblock Adaptive {C}ontext {N}etwork for {S}cene {P}arsing.
\newblock In {\em ICCV}, 2019.

\bibitem[\protect\citeauthoryear{Graham \bgroup \em et al.\egroup
  }{2021}]{graham2021levit}
Benjamin Graham, Alaaeldin El-Nouby, Hugo Touvron, Pierre Stock, Armand Joulin,
  Herv\'e J\'egou, and Matthijs Douze.
\newblock Le{V}i{T}: {A} {V}ision {T}ransformer in {C}onv{N}et's {C}lothing for
  {F}aster {I}nference.
\newblock In {\em ICCV}, October 2021.

\bibitem[\protect\citeauthoryear{Guo \bgroup \em et al.\egroup
  }{2021}]{guo2021pct}
Meng-Hao Guo, Jun-Xiong Cai, Zheng-Ning Liu, Tai-Jiang Mu, Ralph~R Martin, and
  Shi-Min Hu.
\newblock {PCT}: {P}oint {C}loud {T}ransformer.
\newblock {\em Computational Visual Media}, 7(2), 2021.

\bibitem[\protect\citeauthoryear{Guo \bgroup \em et al.\egroup
  }{2022}]{guo2022van}
Meng-Hao Guo, Cheng-Ze Lu, Zheng-Ning Liu, Ming-Ming Cheng, and Shi-Min Hu.
\newblock {Visual Attention Network}.
\newblock {\em arXiv:2202.09741}, 2022.

\bibitem[\protect\citeauthoryear{Han \bgroup \em et al.\egroup
  }{2021}]{han2021transformer}
Kai Han, An~Xiao, Enhua Wu, Jianyuan Guo, Chunjing Xu, and Yunhe Wang.
\newblock Transformer in {T}ransformer.
\newblock In {\em NeurIPS}, 2021.

\bibitem[\protect\citeauthoryear{He \bgroup \em et al.\egroup
  }{2016}]{he2016deep}
Kaiming He, Xiangyu Zhang, Shaoqing Ren, and Jian Sun.
\newblock Deep {R}esidual {L}earning for {I}mage {R}ecognition.
\newblock In {\em CVPR}, 2016.

\bibitem[\protect\citeauthoryear{Heo \bgroup \em et al.\egroup
  }{2021}]{heo2021rethinking}
Byeongho Heo, Sangdoo Yun, Dongyoon Han, Sanghyuk Chun, Junsuk Choe, and
  Seong~Joon Oh.
\newblock Rethinking {S}patial {D}imensions of {V}ision {T}ransformers.
\newblock In {\em ICCV}, 2021.

\bibitem[\protect\citeauthoryear{LeCun \bgroup \em et al.\egroup
  }{2015}]{lecun2015deep}
Yann LeCun, Yoshua Bengio, and Geoffrey Hinton.
\newblock Deep {L}earning.
\newblock {\em Nature}, 521(7553), 2015.

\bibitem[\protect\citeauthoryear{Liu \bgroup \em et al.\egroup
  }{2021}]{liu2021swin}
Ze~Liu, Yutong Lin, Yue Cao, Han Hu, Yixuan Wei, Zheng Zhang, Stephen Lin, and
  Baining Guo.
\newblock Swin {T}ransformer: {H}ierarchical {V}ision {T}ransformer {U}sing
  {S}hifted {W}indows.
\newblock In {\em ICCV}, October 2021.

\bibitem[\protect\citeauthoryear{Liu \bgroup \em et al.\egroup
  }{2022}]{liu2022convnet}
Zhuang Liu, Hanzi Mao, Chao-Yuan Wu, Christoph Feichtenhofer, Trevor Darrell,
  and Saining Xie.
\newblock {A ConvNet for the 2020s}.
\newblock In {\em CVPR}, 2022.

\bibitem[\protect\citeauthoryear{Nagrani \bgroup \em et al.\egroup
  }{2021}]{nagrani2021attention}
Arsha Nagrani, Shan Yang, Anurag Arnab, Aren Jansen, Cordelia Schmid, and Chen
  Sun.
\newblock Attention {B}ottlenecks for {M}ultimodal {F}usion.
\newblock {\em NeurIPS}, 34, 2021.

\bibitem[\protect\citeauthoryear{Naseer \bgroup \em et al.\egroup
  }{2021}]{naseer2021intriguing}
Muzammal Naseer, Kanchana Ranasinghe, Salman Khan, Munawar Hayat, Fahad~Shahbaz
  Khan, and Ming-Hsuan Yang.
\newblock Intriguing {P}roperties of {V}ision {T}ransformers.
\newblock In {\em NeurIPS}, 2021.

\bibitem[\protect\citeauthoryear{OpenMMLab}{2020}]{mmseg2020}
OpenMMLab.
\newblock {{MMS}egmentation}: {O}pen{MML}ab {S}emantic {S}egmentation {T}oolbox
  and {B}enchmark.
\newblock \url{https://github.com/open-mmlab/mmsegmentation}, 2020.
\newblock Accessed: 2023-01-18.

\bibitem[\protect\citeauthoryear{Peng \bgroup \em et al.\egroup
  }{2021}]{peng2021conformer}
Zhiliang Peng, Wei Huang, Shanzhi Gu, Lingxi Xie, Yaowei Wang, Jianbin Jiao,
  and Qixiang Ye.
\newblock Conformer: {L}ocal {F}eatures {C}oupling {G}lobal {R}epresentations
  for {V}isual {R}ecognition.
\newblock In {\em ICCV}, October 2021.

\bibitem[\protect\citeauthoryear{Radosavovic \bgroup \em et al.\egroup
  }{2020}]{radosavovic2020designing}
Ilija Radosavovic, Raj~Prateek Kosaraju, Ross Girshick, Kaiming He, and Piotr
  Doll{\'a}r.
\newblock Designing network design spaces.
\newblock In {\em CVPR}, 2020.

\bibitem[\protect\citeauthoryear{Ranasinghe \bgroup \em et al.\egroup
  }{2022}]{ranasinghe2022self}
Kanchana Ranasinghe, Muzammal Naseer, Salman Khan, Fahad~Shahbaz Khan, and
  Michael~S Ryoo.
\newblock {Self-supervised Video Transformer}.
\newblock In {\em CVPR}, 2022.

\bibitem[\protect\citeauthoryear{Ranftl \bgroup \em et al.\egroup
  }{2021}]{ranftl2021vision}
Ren{\'e} Ranftl, Alexey Bochkovskiy, and Vladlen Koltun.
\newblock Vision {T}ransformers for {D}ense {P}rediction.
\newblock In {\em ICCV}, 2021.

\bibitem[\protect\citeauthoryear{Ryoo \bgroup \em et al.\egroup
  }{2021}]{ryoo2021tokenlearner}
Michael~S Ryoo, AJ~Piergiovanni, Anurag Arnab, Mostafa Dehghani, and Anelia
  Angelova.
\newblock Token{L}earner: {A}daptive {S}pace-{T}ime {T}okenization for
  {V}ideos.
\newblock In {\em NeurIPS}, 2021.

\bibitem[\protect\citeauthoryear{Shang \bgroup \em et al.\egroup
  }{2022}]{shang2021starformer}
Jinghuan Shang, Kumara Kahatapitiya, Xiang Li, and Michael~S Ryoo.
\newblock {StARformer: Transformer with State-Action-Reward Representations}.
\newblock In {\em ECCV}, 2022.

\bibitem[\protect\citeauthoryear{Tan and Le}{2019}]{tan2019efficientnet}
Mingxing Tan and Quoc Le.
\newblock Efficient{N}et: {R}ethinking {M}odel {S}caling for {C}onvolutional
  {N}eural {N}etworks.
\newblock In {\em ICML}. PMLR, 2019.

\bibitem[\protect\citeauthoryear{Tang \bgroup \em et al.\egroup
  }{2021}]{tang2021sparse}
Chuanxin Tang, Yucheng Zhao, Guangting Wang, Chong Luo, Wenxuan Xie, and Wenjun
  Zeng.
\newblock Sparse {MLP} for {I}mage {R}ecognition: {I}s {S}elf-{A}ttention
  {R}eally {N}ecessary?
\newblock {\em arXiv:2109.05422}, 2021.

\bibitem[\protect\citeauthoryear{Tolstikhin \bgroup \em et al.\egroup
  }{2021}]{tolstikhin2021mixer}
Ilya Tolstikhin, Neil Houlsby, Alexander Kolesnikov, Lucas Beyer, Xiaohua Zhai,
  Thomas Unterthiner, Jessica Yung, Andreas~Peter Steiner, Daniel Keysers,
  Jakob Uszkoreit, et~al.
\newblock {MLP}-{M}ixer: {A}n {A}ll-{MLP} {A}rchitecture for {V}ision.
\newblock In {\em NeurIPS}, 2021.

\bibitem[\protect\citeauthoryear{Touvron \bgroup \em et al.\egroup
  }{2021a}]{touvron2021resmlp}
Hugo Touvron, Piotr Bojanowski, Mathilde Caron, Matthieu Cord, Alaaeldin
  El-Nouby, Edouard Grave, Gautier Izacard, Armand Joulin, Gabriel Synnaeve,
  Jakob Verbeek, et~al.
\newblock Res{MLP}: {F}eedforward networks for image classification with
  data-efficient training.
\newblock {\em arXiv:2105.03404}, 2021.

\bibitem[\protect\citeauthoryear{Touvron \bgroup \em et al.\egroup
  }{2021b}]{touvron2021deit}
Hugo Touvron, Matthieu Cord, Matthijs Douze, Francisco Massa, Alexandre
  Sablayrolles, and Herv{\'e} J{\'e}gou.
\newblock Training data-efficient image transformers \& distillation through
  attention.
\newblock In {\em ICML}. PMLR, 2021.

\bibitem[\protect\citeauthoryear{Trockman and
  Kolter}{2022}]{trockman2022patches}
Asher Trockman and J~Zico Kolter.
\newblock Patches are all you need?
\newblock {\em arXiv:2201.09792}, 2022.

\bibitem[\protect\citeauthoryear{Vaswani \bgroup \em et al.\egroup
  }{2017}]{vaswani2017attention}
Ashish Vaswani, Noam Shazeer, Niki Parmar, Jakob Uszkoreit, Llion Jones,
  Aidan~N Gomez, {\L}ukasz Kaiser, and Illia Polosukhin.
\newblock Attention {I}s {A}ll {Y}ou {N}eed.
\newblock In {\em NeurIPS}, 2017.

\bibitem[\protect\citeauthoryear{Wang \bgroup \em et al.\egroup
  }{2018}]{wang2018non}
Xiaolong Wang, Ross Girshick, Abhinav Gupta, and Kaiming He.
\newblock Non-local {N}eural {N}etworks.
\newblock In {\em CVPR}, 2018.

\bibitem[\protect\citeauthoryear{Wang \bgroup \em et al.\egroup
  }{2021}]{wang2021pyramid}
Wenhai Wang, Enze Xie, Xiang Li, Deng-Ping Fan, Kaitao Song, Ding Liang, Tong
  Lu, Ping Luo, and Ling Shao.
\newblock Pyramid {V}ision {T}ransformer: {A} {V}ersatile {B}ackbone for
  {D}ense {P}rediction {W}ithout {C}onvolutions.
\newblock In {\em ICCV}, October 2021.

\bibitem[\protect\citeauthoryear{Wightman}{2019}]{rw2019timm}
Ross Wightman.
\newblock Pytorch {I}mage {M}odels.
\newblock \url{https://github.com/rwightman/pytorch-image-models}, 2019.
\newblock Accessed: 2023-01-18.

\bibitem[\protect\citeauthoryear{Wu \bgroup \em et al.\egroup
  }{2021}]{wu2021cvt}
Haiping Wu, Bin Xiao, Noel Codella, Mengchen Liu, Xiyang Dai, Lu~Yuan, and Lei
  Zhang.
\newblock Cv{T}: {I}ntroducing {C}onvolutions to {V}ision {T}ransformers.
\newblock In {\em ICCV}, October 2021.

\bibitem[\protect\citeauthoryear{Xiao \bgroup \em et al.\egroup
  }{2018}]{xiao2018unified}
Tete Xiao, Yingcheng Liu, Bolei Zhou, Yuning Jiang, and Jian Sun.
\newblock Unified {P}erceptual {P}arsing for {S}cene {U}nderstanding.
\newblock In {\em ECCV}, 2018.

\bibitem[\protect\citeauthoryear{Xiao \bgroup \em et al.\egroup
  }{2021}]{xiao2021early}
Tete Xiao, Piotr Dollar, Mannat Singh, Eric Mintun, Trevor Darrell, and Ross
  Girshick.
\newblock Early {C}onvolutions {H}elp {T}ransformers {S}ee {B}etter.
\newblock In {\em NeurIPS}, 2021.

\bibitem[\protect\citeauthoryear{Xie \bgroup \em et al.\egroup
  }{2017}]{xie2017aggregated}
Saining Xie, Ross Girshick, Piotr Doll{\'a}r, Zhuowen Tu, and Kaiming He.
\newblock Aggregated residual transformations for deep neural networks.
\newblock In {\em CVPR}, 2017.

\bibitem[\protect\citeauthoryear{Xie \bgroup \em et al.\egroup
  }{2021}]{xie2021segformer}
Enze Xie, Wenhai Wang, Zhiding Yu, Anima Anandkumar, Jose~M Alvarez, and Ping
  Luo.
\newblock Seg{F}ormer: {S}imple and {E}fficient {D}esign for {S}emantic
  {S}egmentation with {T}ransformers.
\newblock In {\em NeurIPS}, 2021.

\bibitem[\protect\citeauthoryear{Yang \bgroup \em et al.\egroup
  }{2021a}]{yang2021transformer}
Guanglei Yang, Hao Tang, Mingli Ding, Nicu Sebe, and Elisa Ricci.
\newblock Transformer-{B}ased {A}ttention {N}etworks for {C}ontinuous
  {P}ixel-{W}ise {P}rediction.
\newblock In {\em ICCV}, 2021.

\bibitem[\protect\citeauthoryear{Yang \bgroup \em et al.\egroup
  }{2021b}]{yang2021focal}
Jianwei Yang, Chunyuan Li, Pengchuan Zhang, Xiyang Dai, Bin Xiao, Lu~Yuan, and
  Jianfeng Gao.
\newblock Focal {S}elf-attention for {L}ocal-{G}lobal {I}nteractions in
  {V}ision {T}ransformers.
\newblock In {\em NeurIPS}, 2021.

\bibitem[\protect\citeauthoryear{Yin \bgroup \em et al.\egroup
  }{2020}]{yin2020disentangled}
Minghao Yin, Zhuliang Yao, Yue Cao, Xiu Li, Zheng Zhang, Stephen Lin, and Han
  Hu.
\newblock Disentangled {N}on-{L}ocal {N}eural {N}etworks.
\newblock In {\em ECCV}. Springer, 2020.

\bibitem[\protect\citeauthoryear{Yu \bgroup \em et al.\egroup
  }{2022}]{yu2022metaformer}
Weihao Yu, Mi~Luo, Pan Zhou, Chenyang Si, Yichen Zhou, Xinchao Wang, Jiashi
  Feng, and Shuicheng Yan.
\newblock Metaformer is actually what you need for vision.
\newblock In {\em CVPR}, 2022.

\bibitem[\protect\citeauthoryear{Yuan \bgroup \em et al.\egroup
  }{2021a}]{yuan2021incorporating}
Kun Yuan, Shaopeng Guo, Ziwei Liu, Aojun Zhou, Fengwei Yu, and Wei Wu.
\newblock Incorporating {C}onvolution {D}esigns {I}nto {V}isual {T}ransformers.
\newblock In {\em ICCV}, October 2021.

\bibitem[\protect\citeauthoryear{Yuan \bgroup \em et al.\egroup
  }{2021b}]{yuan2021tokens}
Li~Yuan, Yunpeng Chen, Tao Wang, Weihao Yu, Yujun Shi, Zi-Hang Jiang,
  Francis~E.H. Tay, Jiashi Feng, and Shuicheng Yan.
\newblock Tokens-to-{T}oken {V}i{T}: {T}raining {V}ision {T}ransformers {F}rom
  {S}cratch on {I}mage{N}et.
\newblock In {\em ICCV}, October 2021.

\bibitem[\protect\citeauthoryear{Yue \bgroup \em et al.\egroup
  }{2021}]{yue2021vision}
Xiaoyu Yue, Shuyang Sun, Zhanghui Kuang, Meng Wei, Philip~H.S. Torr, Wayne
  Zhang, and Dahua Lin.
\newblock Vision {T}ransformer {W}ith {P}rogressive {S}ampling.
\newblock In {\em ICCV}, October 2021.

\bibitem[\protect\citeauthoryear{Zhai \bgroup \em et al.\egroup
  }{2021}]{zhai2021attention}
Shuangfei Zhai, Walter Talbott, Nitish Srivastava, Chen Huang, Hanlin Goh,
  Ruixiang Zhang, and Josh Susskind.
\newblock An {A}ttention {F}ree {T}ransformer.
\newblock {\em arXiv:2105.14103}, 2021.

\bibitem[\protect\citeauthoryear{Zhang \bgroup \em et al.\egroup
  }{2021}]{zhang2021aggregating}
Zizhao Zhang, Han Zhang, Long Zhao, Ting Chen, and Tomas Pfister.
\newblock Aggregating {N}ested {T}ransformers.
\newblock {\em arXiv:2105.12723}, 2021.

\bibitem[\protect\citeauthoryear{Zhao \bgroup \em et al.\egroup
  }{2020}]{zhao2020exploring}
Hengshuang Zhao, Jiaya Jia, and Vladlen Koltun.
\newblock Exploring {S}elf-attention for {I}mage {R}ecognition.
\newblock In {\em CVPR}, 2020.

\bibitem[\protect\citeauthoryear{Zhao \bgroup \em et al.\egroup
  }{2021}]{zhao2021point}
Hengshuang Zhao, Li~Jiang, Jiaya Jia, Philip~HS Torr, and Vladlen Koltun.
\newblock Point {T}ransformer.
\newblock In {\em ICCV}, 2021.

\bibitem[\protect\citeauthoryear{Zhou \bgroup \em et al.\egroup
  }{2019}]{zhou2019ade20k}
Bolei Zhou, Hang Zhao, Xavier Puig, Tete Xiao, Sanja Fidler, Adela Barriuso,
  and Antonio Torralba.
\newblock Semantic {U}nderstanding of {S}cenes {T}hrough the {ADE20K}
  {D}ataset.
\newblock {\em IJCV}, 127(3), 2019.

\bibitem[\protect\citeauthoryear{Zhu \bgroup \em et al.\egroup
  }{2020}]{zhu2020deformable}
Xizhou Zhu, Weijie Su, Lewei Lu, Bin Li, Xiaogang Wang, and Jifeng Dai.
\newblock Deformable {DETR}: {D}eformable {T}ransformers for {E}nd-to-{E}nd
  {O}bject {D}etection.
\newblock In {\em ICLR}, 2020.

\end{thebibliography}

\end{document}